%% file: main.tex
\documentclass{article}

\usepackage[preprint]{neurips_2025}
\usepackage{float}
\usepackage{caption}
\usepackage{subcaption}
\usepackage{amsthm}

\usepackage[utf8]{inputenc} 
\usepackage[T1]{fontenc}    
\usepackage{hyperref}       
\usepackage{url}            
\usepackage{booktabs}       
\usepackage{amsfonts}       
\usepackage{nicefrac}       
\usepackage{microtype}      
\usepackage{xcolor}         
\usepackage{graphicx}
\usepackage{amsmath}

\title{Graph-Based Operator Learning from Limited Data on Irregular Domains}

\author{%
Yile Li \\
  Department of Computer Science\\
University of Utah\\
  \texttt{yile.li@utah.edu} \\
  \And
  Shandian Zhe \\
  Department of Computer Science\\
  University of Utah\\
  \texttt{zhe@cs.utah.edu} \\
}

\begin{document}

\maketitle

\input{emacscomm.tex}	





\begin{abstract}
Operator learning seeks to approximate mappings from input functions to output solutions, particularly in the context of partial differential equations (PDEs). While recent advances such as DeepONet and Fourier Neural Operator (FNO) have demonstrated strong performance, they often rely on regular grid discretizations, limiting their applicability to complex or irregular domains. In this work, we propose a \textbf{G}raph-based \textbf{O}perator \textbf{L}earning with \textbf{A}ttention (GOLA) framework that addresses this limitation by constructing graphs from irregularly sampled spatial points and leveraging attention-enhanced Graph Neural Netwoks (GNNs) to model spatial dependencies with global information. To improve the expressive capacity, we introduce a Fourier-based encoder that projects input functions into a frequency space using learnable complex coefficients, allowing for flexible embeddings even with sparse or nonuniform samples. We evaluated our approach across a range of 2D PDEs, including Darcy Flow, Advection, Eikonal, and Nonlinear Diffusion, under varying sampling densities. Our method consistently outperforms baselines, particularly in data-scarce regimes, demonstrating strong generalization and efficiency on irregular domains.

\end{abstract}

\section{Introduction}

Learning mappings between function spaces is a fundamental task in computational physics and scientific machine learning, especially for approximating solution operators of partial differential equations (PDEs). 
 Operator learning offers a paradigm shift by learning the solution operator directly from data, enabling fast, mesh-free predictions across varying input conditions. 
Despite their success, existing operator learning models such as DeepONet \citep{lu2019deeponet} and Fourier Neural Operator (FNO) \citep{li2020fourier} exhibit notable limitations that restrict their applicability in more general settings. A key shortcoming lies in their reliance on regular, uniform grid discretizations. FNO, for instance, requires inputs to be defined on fixed Cartesian grids to leverage fast Fourier transforms efficiently. This assumption limits their flexibility and generalization ability when applied to problems defined on complex geometries, irregular meshes, or unstructured domains, which are common in real-world physical systems. Furthermore, these models often struggle with sparse or non-uniformly sampled data, leading to degraded performance and increased computational cost when adapting to more realistic, heterogeneous scenarios. 

To address these limitations, we propose a \textbf{G}raph-based \textbf{O}perator \textbf{L}earning with \textbf{A}ttention (GOLA) framework that leverages Graph Neural Networks (GNNs) to learn PDE solution operators over irregular spatial domains. By constructing graphs from sampled spatial coordinates and encoding local geometric and functional dependencies through message passing, the model naturally adapts to non-Euclidean geometries.
To enhance global expressivity, we further incorporate attention-based mechanisms that can capture long-range dependencies more effectively and a Fourier-based encoder that projects input functions into a frequency domain using learnable complex-valued bases. 
Our model exhibits superior data efficiency and generalization, achieving smaller prediction errors with fewer training samples and demonstrating robustness under domain shifts.

The main contributions of this work are as follows:
\begin{itemize}
    \item We introduce GOLA, a unified architecture combining spectral encoding and attention-enhanced GNNs for operator learning on irregular domains.
    \item We propose a learnable Fourier encoder that projects input functions into a frequency domain tailored for spatial graphs.
    \item  Through extensive experiments, we demonstrate that GOLA generalizes across PDE types, sample densities, and resolution shifts, achieving state-of-the-art performance in challenging data-scarce regimes.
\end{itemize}

\section{Related Work}



There are many latest research about graph and attention methods in scientific machine learning \citep{pmlr-v235-xiao24c}, \cite{kissas2022learning}, \cite{boulle2024mathematical}, \cite{xu2024equivariant}, \cite{jin2023leveraging},
\cite{cuomo2022physics}
\cite{kovachki2024operator},
\cite{nelsen2024operator},
\cite{batlle2023kernel}.

\textbf{Graph neural networks for scientific machine learning.} \cite{battaglia2018relational} applies shared functions over nodes and edges, captures relational inductive biases and generalizes across different physical scenarios.
\cite{barsinai2019learning} learns data-driven discretization schemes for solving PDEs by training a neural network to predict spatial derivatives directly from local stencils. By replacing hand-crafted finite difference rules with learned operators, it adapts discretizations to the underlying data for improved accuracy and generalization.
\cite{sanchezgonzalez2020learning} predicts future physical states by performing message passing over the mesh graph, capturing both local and global dynamics without relying on explicit numerical solvers.
 Graph Kernel Networks (GKNs) \citep{li2020neural} directly approximates continuous mappings between infinite-dimensional function spaces by utilizing graph kernel convolution layers.
PDE-GCN \citep{wang2022pdegcn} represents partial differential equations on arbitrary graphs by combining spectral graph convolution with PDE-specific inductive biases. It learns to predict physical dynamics directly on graph-structured domains, enabling generalization across varying geometries and discretizations.
The Message Passing Neural PDE Solver \citep{brandstetter2022message} 
 formulates spatiotemporal PDE dynamics by applying learned message passing updates on graph representations of the solution domain.
Physics-Informed Transformer (PIT) \citep{zhang2024attention} embeds physical priors into the Transformer architecture to model PDE surrogate solutions. It leverages self-attention to capture long-range dependencies and integrates PDE residuals as soft constraints during training to improve generalization.
GraphCast \citep{lam2024graphcast} learns the Earth’s atmosphere as a spatiotemporal graph and uses a graph neural network to iteratively forecast future weather states based on past observations. It performs message passing over the graph to capture spatial correlations and temporal dynamics, enabling accurate medium-range forecasts.


\textbf{Attention-based methods for scientific machine learning.}
U-Netformer \citep{liu2022unetformer} proposes a hybrid neural architecture that combines the U-Net's hierarchical encoder-decoder structure with transformer-based attention modules to capture both local and global dependencies in PDE solution spaces. 
Tokenformer \citep{zhou2023tokenformer} reformulates PDE solving as a token mixing problem by representing input fields as tokens and applying self-attention across them to model spatial correlations.
Tokenized Neural Operators (TNO) \citep{jiang2023tokenized} framework reformulates operator learning as a token-wise attention problem, where spatial inputs are encoded into tokens and processed using Transformer-style architectures. It employs a combination of axial attention and cross-attention to efficiently capture both local and global dependencies across irregular or structured domains.
Attention-Based Fourier Neural Operator (AFNO) \citep{zhang2024attention} enhances the standard FNO by integrating self-attention layers after each Fourier convolution block to better capture local dependencies and long-range interactions. This hybrid architecture improves the model’s ability to approximate complex solution operators for parametric PDEs by combining spectral global context with spatial attention mechanisms.

Our proposed GOLA combines the local relational strengths of attention-enhanced GNNs and the global spectral capabilities of Fourier-based encoding. This hybrid approach has shown notable improvements in generalization and data efficiency, particularly under challenging data-scarce conditions on irregular domains.

\section{Methodology}

\subsection{Problem Formulation}


Consider the general form of a PDE
\begin{equation}
    \mathcal{N}[u](\mathbf{x}) = f(\mathbf{x}), \quad \mathbf{x} \in \Omega \times [0, \infty)
\end{equation}
where $\mathbf{x}$ denotes a compact representation of the spatial and temporal coordinates, $\Omega$ is the spatial domain, and $[0, \infty)$ is the temporal domain. $\mathcal{N}$ is a differential operator, $u(\mathbf{x})$ is the unknown solution, and $f(\mathbf{x})$ is a given source term.
The objective is to learn the solution operator $\mathcal{G}: \mathcal{F} \to \mathcal{U}$, where $\mathcal{F}$ and $\mathcal{U}$ are Banach spaces.
We assume access to a training dataset $\mathcal{D} = \{ (f_n, u_n) \}_{n=1}^N$, consisting of multiple input-output function pairs, where each $f_n(\cdot)$ and $u_n(\cdot)$ is represented by discrete samples over a finite set of points.

While existing approaches such as DeepONet and FNO have demonstrated strong performance, they typically rely on structured, grid-based discretizations of the domain. This assumption limits their applicability to unstructured meshes, complex geometries, and adaptively sampled domains.
To overcome this limitation, we employ GNNs for operator learning by representing the domain as a graph. This allows for modeling on arbitrary domains and sampling patterns. 
Once trained, the operator learning model can efficiently predict the solution $u$ for a new instance of the input $f$ at random locations.

\begin{figure}[H]
    \centering
    \includegraphics[width=0.8\textwidth]{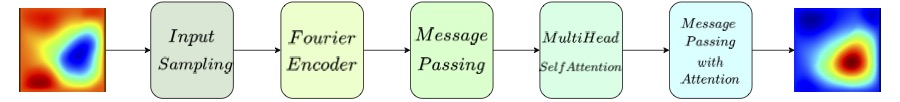}  
    \label{fig:my_pdf}
\end{figure}

\begin{figure}[H]
    \centering
    \includegraphics[width=0.8\textwidth]{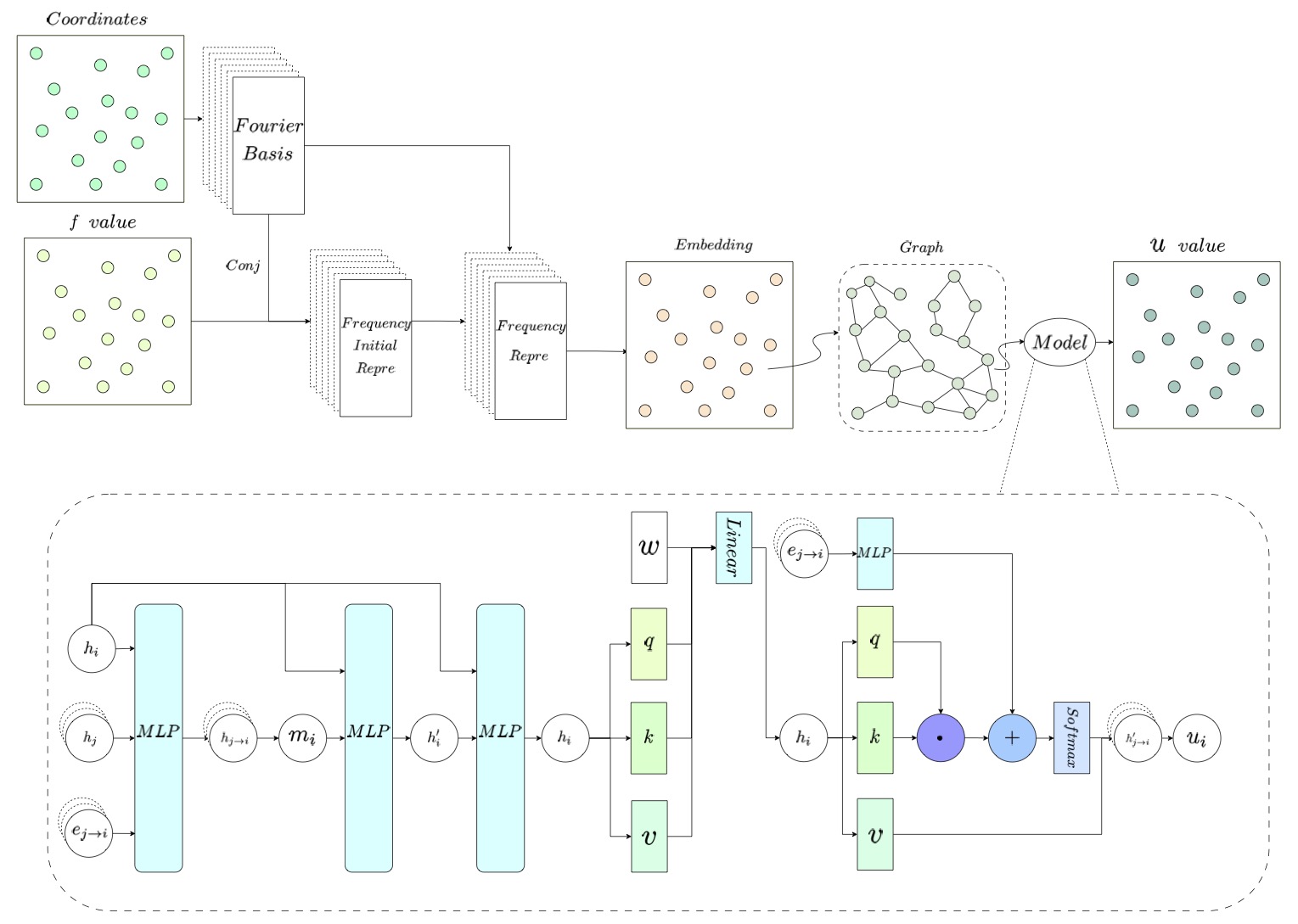}  
    \caption{\small \textit{GOLA Model}}
\end{figure}

\subsection{Graph Construction}

To represent PDE solutions over irregular domains, we begin by randomly sampling a subset of points $\{x_i\}^N_{i=1}$ from a uniform grid in 2D space. We then construct a graph $G = (V, E)$ with nodes $V=\{x_i\}$ and edges $E$ determined by a radius $r$. Edges are created based on spatial proximity. Two nodes are connected if the Euclidean distance between them is less than a threshold $r$ such that 
  $(i,j) \in E$ if and only if $\| x_i - x_j \|_2 \leq r$.
Each edge $(i, j)$ carries edge attributes $e_{ij}$ that encode both geometric and feature-based information, such as the relative coordinates and function values at nodes $i$ and $j$ such that $e_{ij} = \mathbin\Vert (x_i, x_j, f(x_i), f(x_j))$,
where $\mathbin\Vert$ is the concatenation operation. This graph-based representation allows us to model unstructured spatial domains and enables message passing among nonuniform samples.

\subsection{Fourier Encoder}

We define a set of learnable frequencies
$   \left\{ \omega_m \in \mathbb{R}^2 \,\middle|\, m = 1, \ldots, M \right\}$.

For any coordinate $ x \in \mathbb{R}^2 $, the $m$-th basis function is given by the complex exponential
\begin{align}
    \varphi_m(x) = e^{2\pi i \langle \omega_m, x \rangle}
\end{align}

where $\langle \cdot, \cdot \rangle$ denotes the standard Euclidean inner product, and $i$ is the imaginary unit.

At the discrete level, for a batch of $B$ samples and $N$ points per sample, the basis matrix is defined as
\begin{align}
    \Phi \in \mathbb{C}^{B \times N \times M}, \quad \Phi_{b,i,m} = e^{2\pi i \langle \omega_m, x_i^{(b)} \rangle}
\end{align}
where $x_i^{(b)}$ denotes the $i$-th coordinate point in the $b$-th batch sample.

Given the input $f \in \mathbb{R}^{B \times C_{\text{in}} \times N}$ sampled at points $\{ x_i \}$, we first project onto the Fourier basis.
We compute the Fourier coefficients by
\begin{align}
    \hat{u}_{b,c,m} = \frac{1}{N} \sum_{i=1}^{N} f_{b,c,i} \, \overline{\varphi_m\left(x_i^{(b)}\right)}
\end{align}
where $\overline{(\cdot)}$ denotes complex conjugation.

We introduce a learnable set of complex Fourier coefficients $   W \in \mathbb{C}^{C_{\text{in}} \times C_{\text{out}} \times M}$.
The spectral filtering operation is
\begin{align}
    \hat{v}_{b,o,m} = \sum_{c=1}^{C_{\text{in}}} \hat{u}_{b,c,m} \, W_{c,o,m}
\end{align}

We reconstruct the output in the physical domain by applying the inverse transform
\begin{align}
    v_{b,o,i} = \sum_{m=1}^{M} \hat{v}_{b,o,m} \, \varphi_m\left(x_i^{(b)}\right)
\end{align}

Since $v$ is complex-valued, we only take its real part for the output as $h = \text{Re}(v) \in \mathbb{R}^{B \times C_{\text{out}} \times N}$.
The output $h$ serves as the input node features for the downstream GNN model.

\subsection{Message Passing}

Given a node $i \in V$ and its set of neighbors $\mathcal{N}(i)$, the pre-processed messages $\{ m_{ij} \}_{j \in \mathcal{N}(i)}$ are first computed using a learnable neural network $g_{\Theta}$ as
\begin{align}
  m_{ij} = g_{\Theta}(h_i, h_j, e_{ij})  
\end{align}
where $h_i$ and $h_j$ are node features, and $e_{ij}$ denotes edge attributes.

Then we aggregate message from neighbors such that
\begin{align}
    \hat{m} = \mathbin\Vert (\frac{1}{|\mathcal{N}(i)|} \sum_{j \in \mathcal{N}(i)} m_{ij}, \max_{j \in \mathcal{N}(i)} m_{ij}, \min_{j \in \mathcal{N}(i)} m_{ij},  \sqrt{ \frac{1}{|\mathcal{N}(i)|} \sum_{j \in \mathcal{N}(i)} (m_{ij} - \frac{1}{|\mathcal{N}(i)|} \sum_{j \in \mathcal{N}(i)} m_{ij})^2 })
\end{align}

This concatenated feature vector is processed by a post-aggregation neural network \( \gamma_{\Theta} \) to produce the updated node representation by
\begin{align}
  h_i' = \gamma_{\Theta}\left( h_i, \hat{m} \right)
\end{align}
The updated node representation is passed through additional MLP layers with residual connections to enhance expressiveness.

\subsection{Multi-Head Self-Attention}

We employ $H$ independent attention heads. For each head $h$,
the query, key and value functions are computed as linear projections
\begin{align}
q^{(h)}(x) = W_q h'(x), \quad
k^{(h)}(y) = W_k h'(y), \quad
v^{(h)}(y) = W_v h'(y)
\end{align}
where $W_q, W_k, W_v \in \mathbb{R}^{d_h \times C_{\text{out}}}$, $q^{(h)}(x), k^{(h)}(y), v^{(h)}(y) \in \mathbb{R}^{d_h}$ are learned head-specific features, and $d_h$ is the dimension per attention head.

Before computing attention, the keys and values are normalized
\begin{align}
\tilde{k}^{(h)}(y) = \text{Norm}(k^{(h)}(y)), \quad
\tilde{v}^{(h)}(y) = \text{Norm}(v^{(h)}(y))
\end{align}
where $\text{Norm}(\cdot)$ denotes instance normalization.

We compute
\begin{align}
G_h = \sum_{j=1}^N \tilde{k}^{(h)}(y_j)^\top \tilde{v}^{(h)}(y_j) w(y_j),
\quad
(\mathcal{K}_h h')(x_i) = q^{(h)}(x_i) G_h
\end{align}


The outputs are concatenated and projected to the output space by
\begin{align}
  (\mathcal{K}h')(x_i) = \mathbin\Vert \left( (\mathcal{K}_1 h')(x_i), \ldots, (\mathcal{K}_H h')(x_i) \right),
\quad
\hat{h}(x_i) = W_{\text{out}} (\mathcal{K}h')(x_i)
\end{align}
where where $G_h \in \mathbb{R}^{d_h \times d_h}$, $w$ is calculated by the number of points, $W_{\text{out}} \in \mathbb{R}^{C_{\text{out}} \times (C_{\text{out}} \cdot H)}$.

The result is then passed through a linear projection layer to update the node features.

\subsection{Message Passing with Attention}
We update node features and add a skip connection by
\begin{align}
  \hat{h}_i' = W_1 \hat{h}_i + \sum_{j \in \mathcal{N}(i)} \alpha_{ij} \left( W_2 \hat{h}_j + W_3 e_{ij} \right),  
  \quad
    \hat{h}_i' = \hat{h}_i' + W_s \hat{h}_i
\end{align}

The attention weights $\alpha_{ij}$ are computed using a scaled dot-product attention mechanism by
\begin{align}
    \alpha_{ij} = \text{softmax}_j \left( \frac{ \left( W_4 \hat{h}_i \right)^\top \left( W_5 \hat{h}_j + W_3 e_{ij} \right) }{ \sqrt{d} } \right)
\end{align}
where $d$ is the dimensionality of the head, and the softmax is applied over the set of neighbors $j \in \mathcal{N}(i)$.
Then we add a linear projection to produce the predicted solution $\hat{u}$.

\subsection{Training}

The model is trained to minimize the relative $L_2$ error between predicted and true solutions by
\begin{align}
    \mathcal{L}_2(\theta) = \frac{ \| u - \mathcal{G}_\theta(f) \|_{L^2(\Omega)} }{ \| u \|_{L^2(\Omega)} }
\end{align}

\section{Theoretical Analysis}
\label{theory}
Following the universal approximation theorem for operators \citep{lu2019deeponet}, neural operator architectures can approximate any continuous operator $\mathcal{G}$ between Banach spaces when provided with sufficient capacity. 

\paragraph{Proposition.} Let $\mathcal{G}: \mathcal{F} \rightarrow \mathcal{U}$ be a continuous nonlinear operator between separable Banach spaces. Then, under sufficient model capacity, the GOLA architecture $\mathcal{G}_\theta$ can approximate $\mathcal{G}$ arbitrarily well in the $L^2(\Omega)$ norm over a compact domain $\Omega$, i.e.,
$\sup_{f \in \mathcal{F}_\delta} \| \mathcal{G}(f) - \mathcal{G}_\theta(f) \|_{L^2(\Omega)} < \epsilon$,
for any $\epsilon > 0$ and compact subset $\mathcal{F}_\delta \subset \mathcal{F}$.

\textit{Proof}. Given a function $f \in \mathcal{F} \subset L^2(\Omega)$, we sample it at $N$ spatial locations $\{x_i\}_{i=1}^N \subset \Omega$ to obtain a discrete representation $f_N = (f(x_1), \ldots, f(x_N)) \in \mathbb{R}^N.$
Since $\Omega$ is compact, by increasing $N$ the point cloud $\{x_i\}$ becomes dense in $\Omega$. Thus, $f_N$ can approximate $f$ arbitrarily well in $L^2(\Omega)$ norm via interpolation over the sampling set.

Define a set of complex Fourier basis functions $\{\phi_m(x) = e^{2\pi i \langle \omega_m, x \rangle}\}_{m=1}^M$. The Fourier basis is complete in $L^2(\Omega)$, so for any $f \in \mathcal{F}$ and $\delta > 0$, there exists $M$ such that
\begin{align*}
    \left\| f(x) - \sum_{m=1}^M \hat{f}_m \phi_m(x) \right\|_{L^2(\Omega)} < \delta.
\end{align*}

This guarantees that the learnable Fourier encoder in GOLA can approximate the functional input $f$ to arbitrary precision.

Construct a graph $G=(V, E)$ with node set $V=\{x_i\}_{i=1}^N$, where edges encode local spatial relationships. 
According to universal approximation results for GNNs \citep{xu2019powerful}, \citep{morris2019weisfeiler}, for any continuous function defined on graphs, a GNN with sufficient depth and width can approximate it arbitrarily well. Thus, the GNN decoder can approximate the mapping from input features to solution values
\begin{align*}
    (f(x_1), \dots, f(x_N)) \mapsto (\mathcal{G}(f)(x_1), \dots, \mathcal{G}(f)(x_N))
\end{align*}

Let $\mathcal{T}_N$ denote the sampling operator, $\mathcal{F}_\theta$ the Fourier encoder, and $\mathcal{D}_\theta$ the GNN decoder. Then the GOLA operator can be written as
\begin{align*}
    \mathcal{G}_\theta = \mathcal{D}_\theta \circ \mathcal{F}_\theta \circ \mathcal{T}_N
\end{align*}
Each component is continuous and approximates its target arbitrarily well. Since composition of continuous approximations preserves continuity, and $\mathcal{F}_\delta$ is compact, the total approximation error can be made less than any $\varepsilon > 0$ by choosing $N$, $M$, and model capacity large enough such that
\begin{align*}
    \sup_{f \in \mathcal{F}_\delta} \| \mathcal{G}(f) - \mathcal{G}_\theta(f) \|_{L^2(\Omega)} < \varepsilon
\end{align*}

\section{Experiments}
We evaluate the proposed model GOLA on four 2D PDE benchmarks including Darcy Flow, Nonlinear Diffusion, Eikonal, and Advection.
For each dataset, we simulate training data with 5, 10, 20, 30, 40, 50, 80, 100 samples and use 100 examples for testing.
To construct graphs, we randomly sample 20, 30, 40, 50, 60, 70, 80, 90, 100, 200, 300, 400, 500, 600, 700, 800, 900, 1000 points from a uniform \( 128 \times 128 \) grid over the domain \([0, 1] \times [0, 1]\). The sampled points define the nodes of the graph.
Our model learns to approximate the solution operator from these irregularly sampled inputs. We aim to test generalization under both limited data and resolution changes.
We compare against the following baselines including DeepONet, FNO, Graph Convolutional Network (GCN), and Graph Kernel Network (GKN).
We test DeepONet, FNO, GCN on standard grid data, and test GKN and our model GOLA on 1000 sample data randomly selected from grid data.

\begin{figure}[H]
    \centering
    \begin{subfigure}[b]{0.32\textwidth}
        \includegraphics[width=\textwidth]{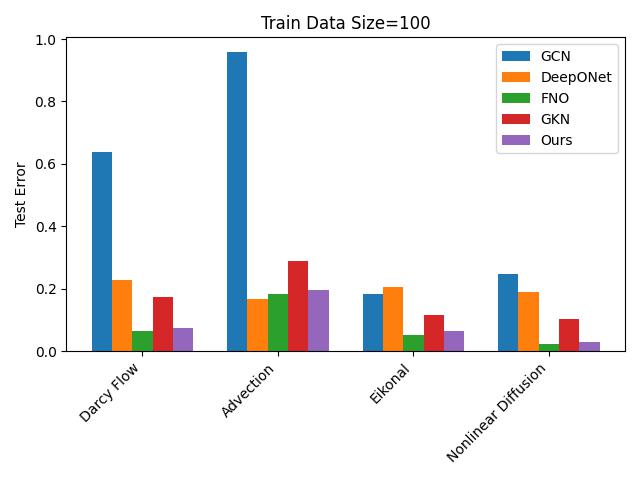}
    \end{subfigure}
    \begin{subfigure}[b]{0.32\textwidth}
        \includegraphics[width=\textwidth]{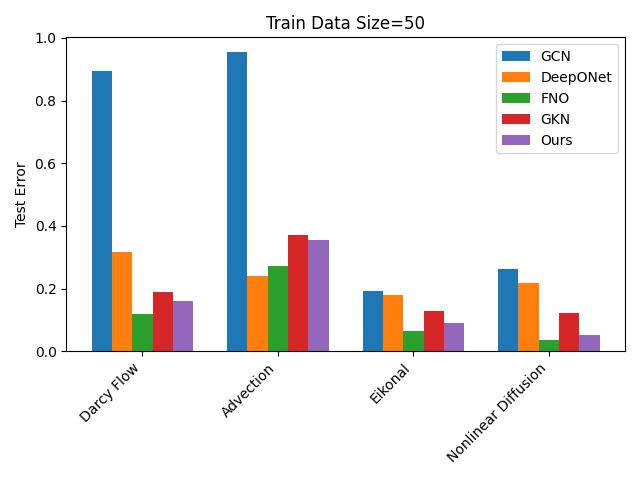}
    \end{subfigure}
    \begin{subfigure}[b]{0.32\textwidth}
        \includegraphics[width=\textwidth]{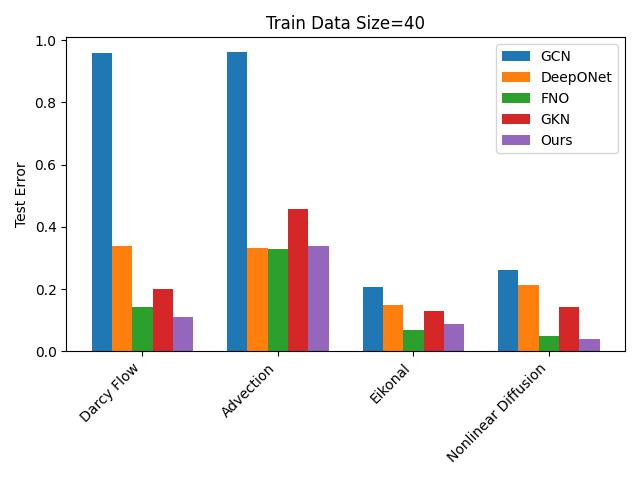}
    \end{subfigure}
    
    \begin{subfigure}[b]{0.32\textwidth}
        \includegraphics[width=\textwidth]{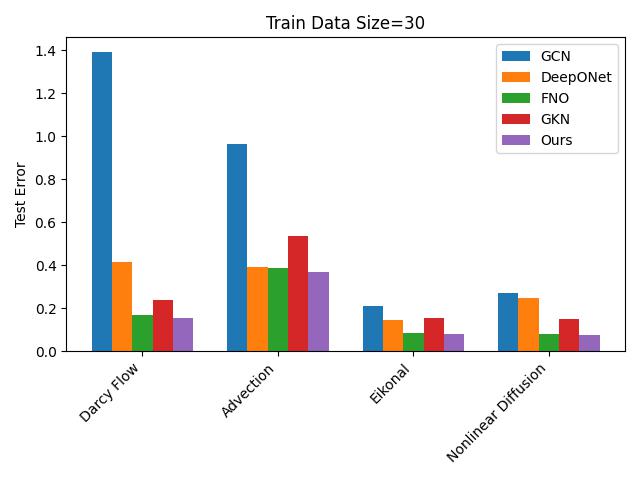}
    \end{subfigure}
    \begin{subfigure}[b]{0.32\textwidth}
        \includegraphics[width=\textwidth]{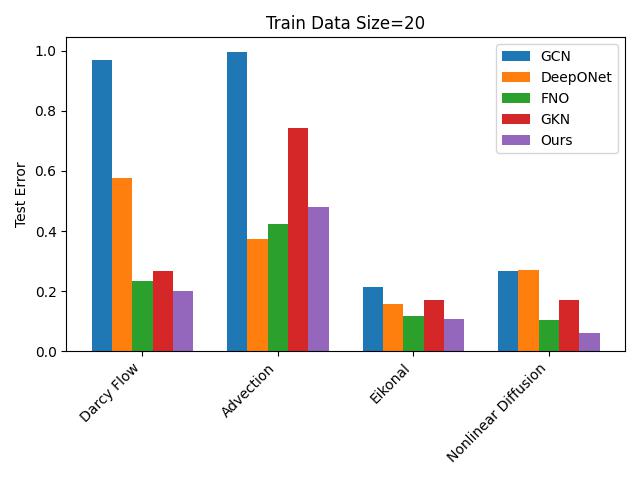}
    \end{subfigure}
    \begin{subfigure}[b]{0.32\textwidth}
        \includegraphics[width=\textwidth]{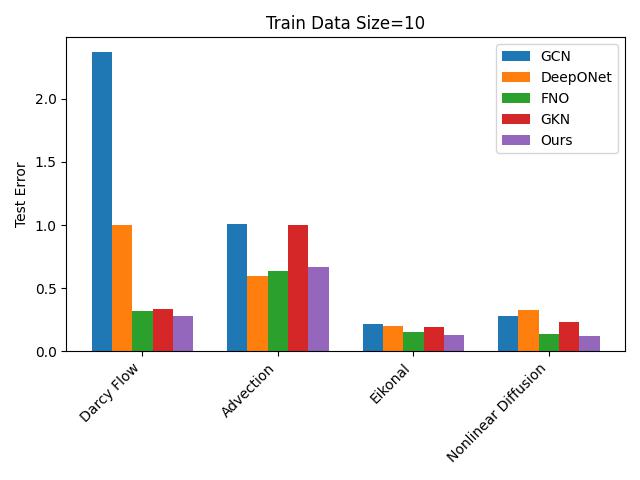}
    \end{subfigure}
    
    \caption{\small \textit{Test errors across PDE benchmarks with varying training data sizes}}
    \label{fig:test-errors-train-all}
\end{figure}

\begin{figure}[htbp]
    \centering

    \begin{subfigure}[b]{0.32\textwidth}
        \includegraphics[width=\textwidth]{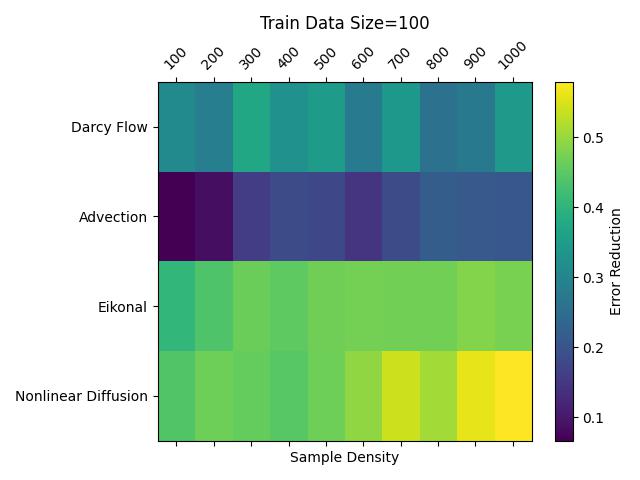}
    \end{subfigure}
    \hfill
    \begin{subfigure}[b]{0.32\textwidth}
        \includegraphics[width=\textwidth]{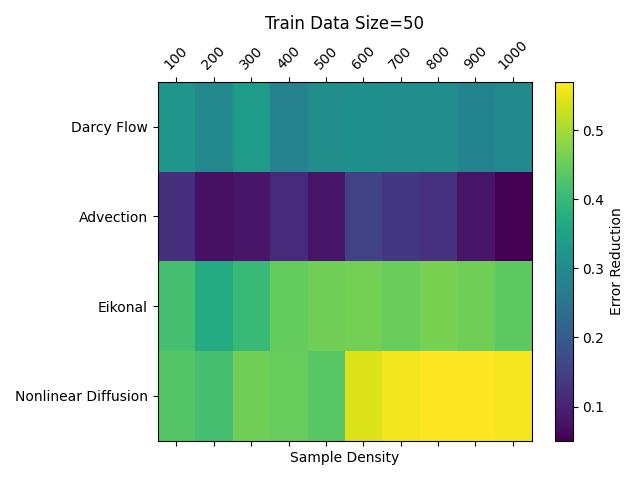}
    \end{subfigure}
    \hfill
    \begin{subfigure}[b]{0.32\textwidth}
        \includegraphics[width=\textwidth]{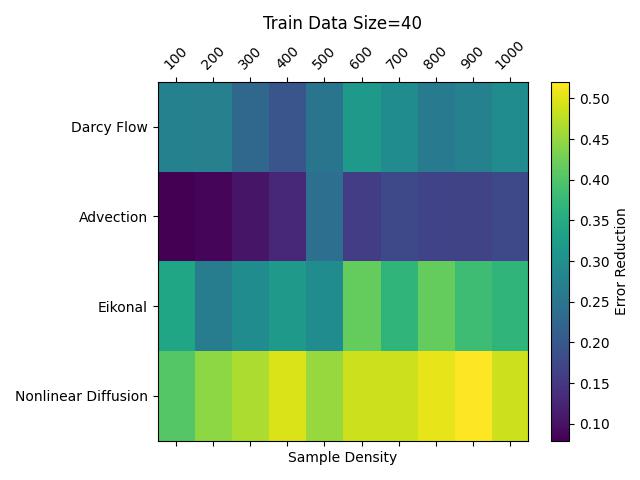}
    \end{subfigure}

    \vskip\baselineskip

    \begin{subfigure}[b]{0.32\textwidth}
        \includegraphics[width=\textwidth]{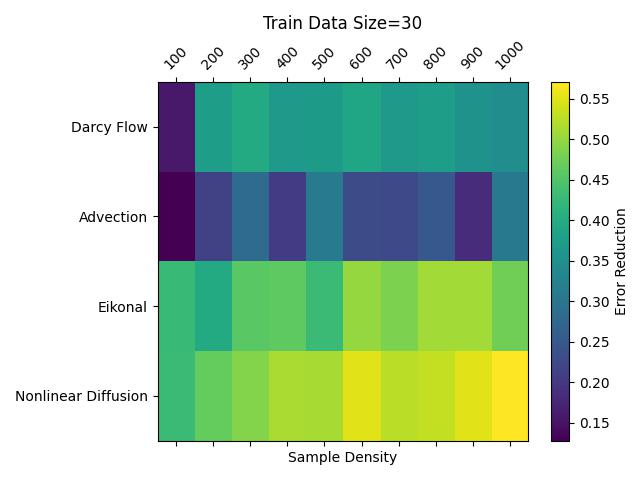}
    \end{subfigure}
    \hfill
    \begin{subfigure}[b]{0.32\textwidth}
        \includegraphics[width=\textwidth]{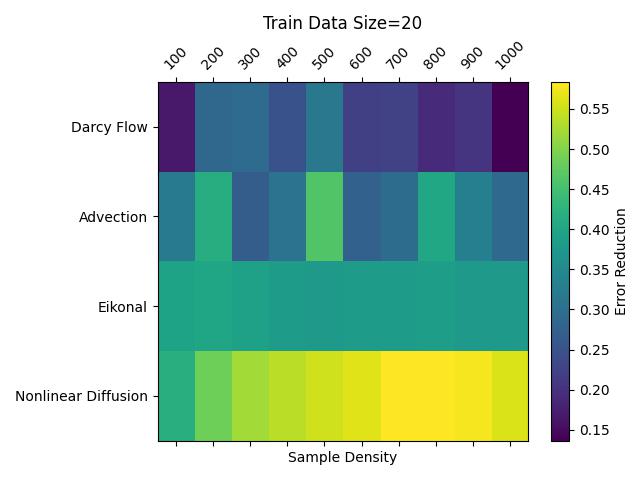}
    \end{subfigure}
    \hfill
    \begin{subfigure}[b]{0.32\textwidth}
        \includegraphics[width=\textwidth]{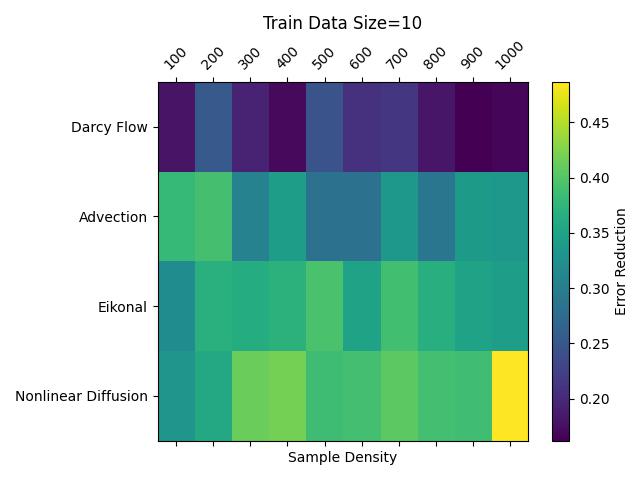}
    \end{subfigure}
    \caption{\small \textit{Error reduction heatmaps across training data sizes and sample densities for PDE Benchmarks}}
    \label{fig:heatmaps_train_sizes}
\end{figure}

\textbf{Comparison with baselines.} From Table \ref{fig:comparison_baseline}, we test on 30 training data, our model consistently achieves the lowest test error across all four benchmarks. For example, in Darcy Flow, GCN whose relative $L_2$ error is 1.3902 performs the worst, suggesting graph-only models may lack the expressiveness needed here. FNO and GKN improve significantly whose relative $L_2$ error are 0.1678 and 0.2375 respectively, showing the benefit of spectral and kernel-based methods. GOLA further improves performance with a relative $L_2$ error 0.1554. From Table \ref{tab:error_reduction}, we use 100 training data, and 1000 sample points. It shows that our method GOLA outperforms GKN across all four datasets Darcy Flow, Advection, Eikonal, Nonlinear Diffusion with error reduction $34.38\%, 20.65\%, 47.69\%, 57.95\%$ respectively.
In Figure \ref{fig:test-errors-train-all}, we display test error comparisons across four PDE benchmarks for five different models. GCN struggles across all PDEs, especially as data decreases. DeepONet and FNO show stronger performance, particularly with more data. GKN and GOLA are more data-efficient. And GOLA generalizes best under limited training samples, across all four PDE types.
In Figure \ref{fig:heatmaps_train_sizes}, we present heatmaps of error reduction across different PDE benchmarks under varying training data sizes and sample densities. Nonlinear Diffusion consistently shows the highest error reduction across all training sizes and densities. And it becomes more prominent at high sample densities even under very small training sizes such as 10.

\begin{table}[H]
\centering
\caption{\small \textit{Test errors for different models trained on 30 training data samples across various PDE benchmarks}}
\label{fig:comparison_baseline}
\small
\begin{tabular}{lccccc}
\toprule
\textit{Dataset} & GCN & DeepONet & FNO & GKN & Ours \\
\midrule
Darcy Flow & 1.3902 & 0.4163 & 0.1678 & 0.2375 & 0.1554 \\
Advection & 0.9620 & 0.3925 & 0.3873 & 0.5353 & 0.3707 \\
Eikonal & 0.2120 & 0.1442 & 0.0874 & 0.1530 & 0.0803 \\
Nonlinear Diffusion & 0.2703 & 0.2468 & 0.0793 & 0.1520 & 0.0747 \\
\bottomrule
\end{tabular}
\end{table}

\begin{table}[H]
\centering
\caption{\small \textit{Comparison of GKN and GOLA with sample density=1000, train data size=100}}
\label{tab:error_reduction}
\small
\begin{tabular}{lccc}
\toprule
\textit{Dataset} & GKN & Ours & Error Reduction \\
\midrule
Darcy Flow & 0.1748 & 0.1147 & 34.38\% \\
Advection & 0.2886 & 0.2290 & 20.65\% \\
Eikonal & 0.1168 & 0.0611 & 47.69\% \\
Nonlinear Diffusion & 0.1044 & 0.0439 & 57.95\% \\
\bottomrule
\end{tabular}
\end{table}

\textbf{Generalization across sample densities.}
From Table \ref{tab:Test_errors_test_sampling_densities}, we use 100 training data, and choose three types of sampling densities 20, 500, 1000 which represent small, medium and high sample densities. We can conclude that increasing sample density leads to smaller test error.
\begin{table}[H]
        \centering
    \caption{ \small \textit{Test errors for small, medium, and high sampling densities with training data size=100}}
    \small
    \label{tab:Test_errors_test_sampling_densities}
    \begin{tabular}{lccc}
        \toprule
        \text{\textit{Sample Density}} & \text{20} & \text{500} & \text{1000} \\
        \midrule
        Darcy Flow       & 0.4269 & 0.1433 & 0.1147 \\
        Advection  & 0.3980 & 0.2462 & 0.2290 \\
        Eikonal         & 0.1236 & 0.0671 & 0.0611 \\
        Nonlinear Diffusion  & 0.2030 & 0.0637 & 0.0439 \\
        \bottomrule
    \end{tabular}
\end{table}

\begin{figure}[H]
    \centering
    \begin{minipage}[b]{0.48\linewidth}
        \centering
    \includegraphics[width=\linewidth]{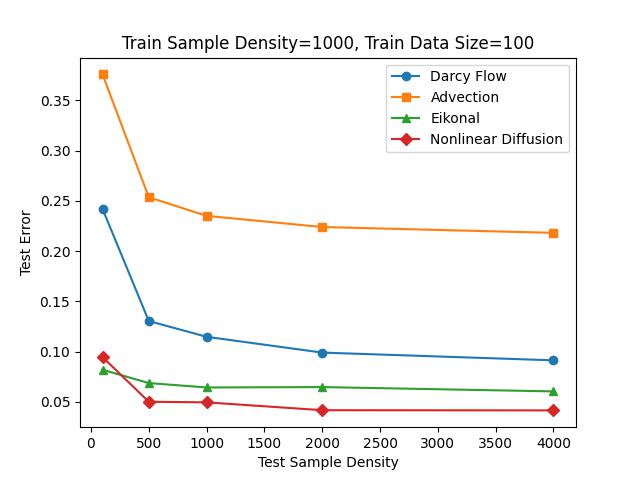}
    \caption{\small \textit{Test error trend with test sample density}}
    \label{fig:test-error-test}
    \end{minipage}
    \hfill
    \begin{minipage}[b]{0.48\linewidth}
        \centering
    \includegraphics[width=\linewidth]{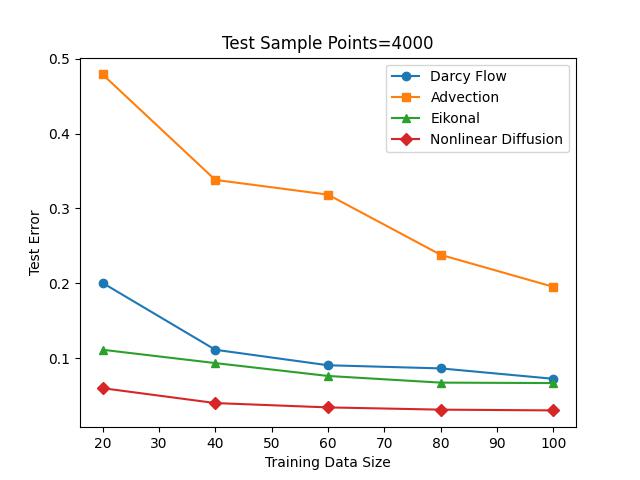}
    \caption{\small \textit{Test error trend with train data size}}
    \label{fig:test-error-train}
    \end{minipage}
\end{figure}

\textbf{Resolution generalization.} From Table \ref{tab:vary_test_pts} and Figure \ref{fig:test-error-test}, we use 100 training data and sample 1000 training sample points, then we test the relative $L_2$ error in different test sample densities 100, 500, 1000, 2000, 4000. It shows that the test error decreases as test sample density increases, suggesting that the model generalizes better when evaluated on finer-resolution grids even though it is trained on a coarser sampling.

\begin{table}[H]
    \centering
    \caption{ \small\textit{Test errors for different test sampling densities with training sample density=1000}}
    \label{tab:vary_test_pts}
    \small
    \begin{tabular}{lccccc}
        \toprule
        \text{\textit{Test Sample Density}}                      & 100     & 500     & 1000    & 2000 & 4000    \\
        \midrule
        Darcy Flow            & 0.2419  & 0.1304  & 0.1147  & 0.0990  & 0.0913  \\
        Advection             & 0.3757  & 0.2535  & 0.2290  & 0.2239  & 0.2181  \\
        Eikonal               & 0.0819  & 0.0687  & 0.0611  & 0.0610  & 0.0604  \\
        Nonlinear Diffusion   & 0.0945  & 0.0501  & 0.0439  & 0.0417  & 0.0415  \\
        \bottomrule
    \end{tabular}
\end{table}

\begin{figure}[htbp]
    \centering

    \begin{subfigure}[b]{0.23\textwidth}
        \includegraphics[width=\textwidth]{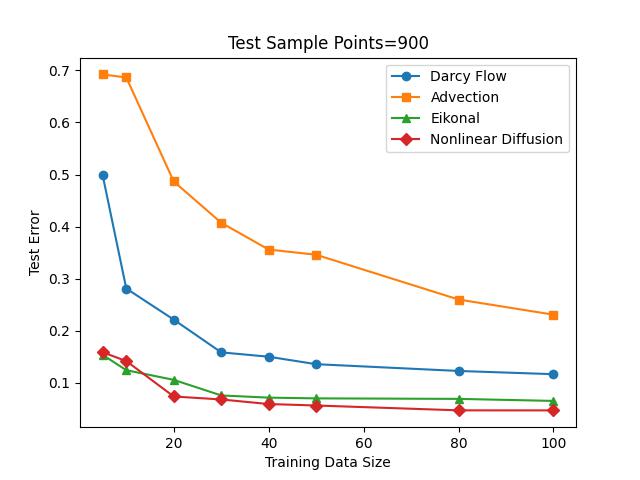}
    \end{subfigure}
    \hfill
    \begin{subfigure}[b]{0.23\textwidth}
        \includegraphics[width=\textwidth]{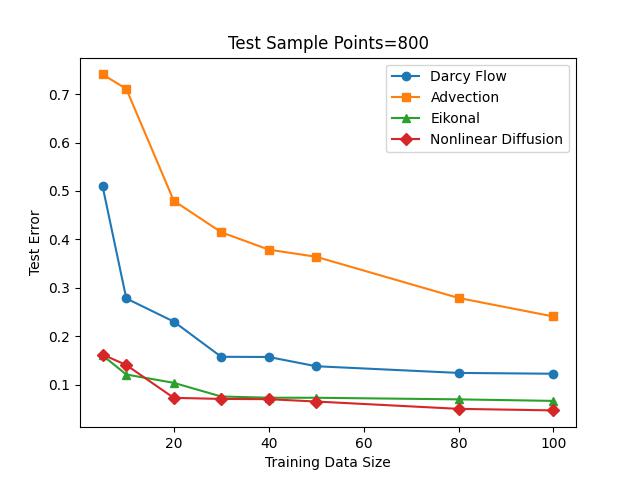}
    \end{subfigure}
    \hfill
    \begin{subfigure}[b]{0.23\textwidth}
        \includegraphics[width=\textwidth]{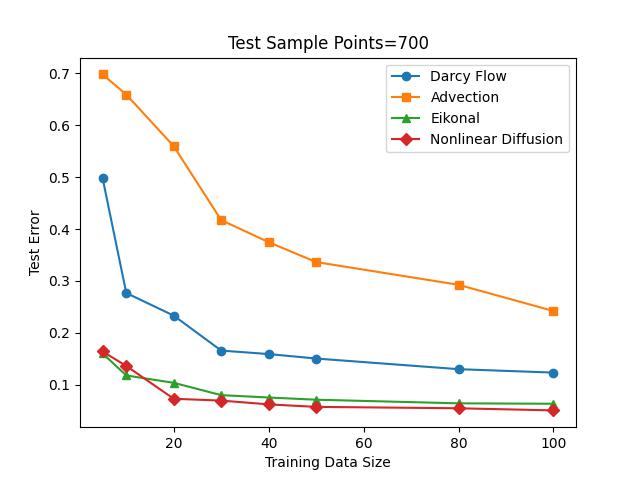}
    \end{subfigure}
    \hfill
    \begin{subfigure}[b]{0.23\textwidth}
        \includegraphics[width=\textwidth]{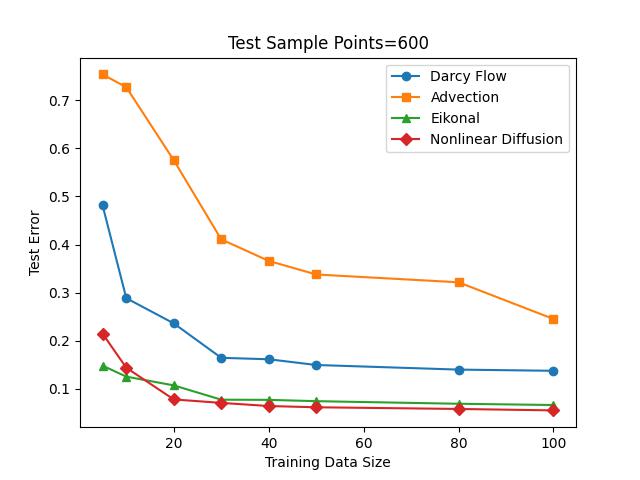}
    \end{subfigure}

    \vskip\baselineskip

    \begin{subfigure}[b]{0.23\textwidth}
        \includegraphics[width=\textwidth]{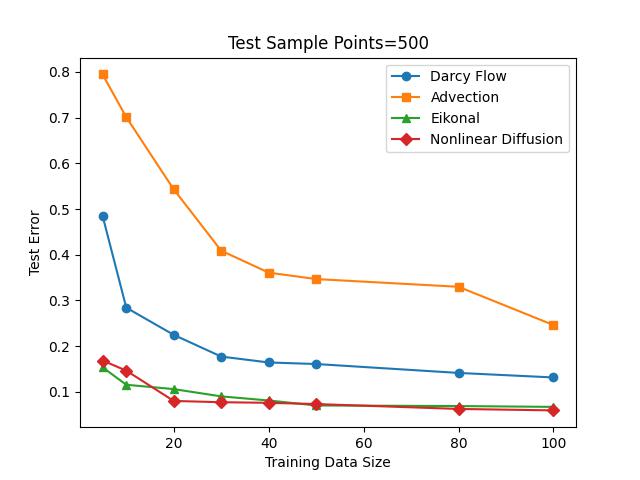}
    \end{subfigure}
    \hfill
    \begin{subfigure}[b]{0.23\textwidth}
        \includegraphics[width=\textwidth]{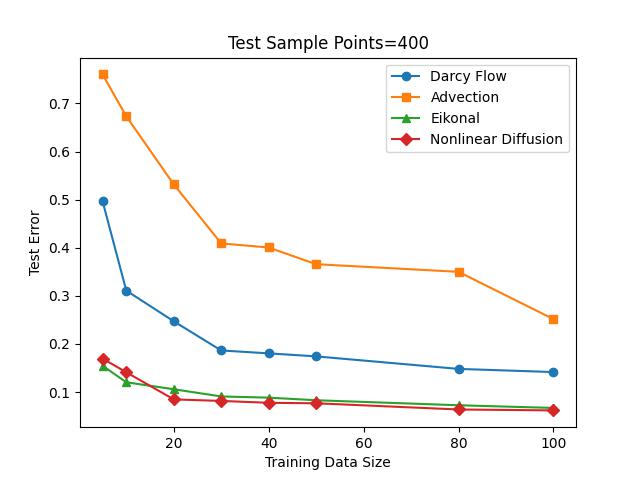}
    \end{subfigure}
    \hfill
    \begin{subfigure}[b]{0.23\textwidth}
        \includegraphics[width=\textwidth]{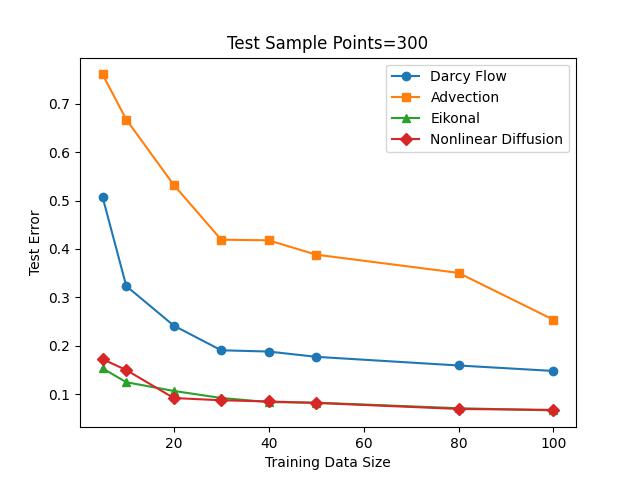}
    \end{subfigure}
    \hfill
    \begin{subfigure}[b]{0.23\textwidth}
        \includegraphics[width=\textwidth]{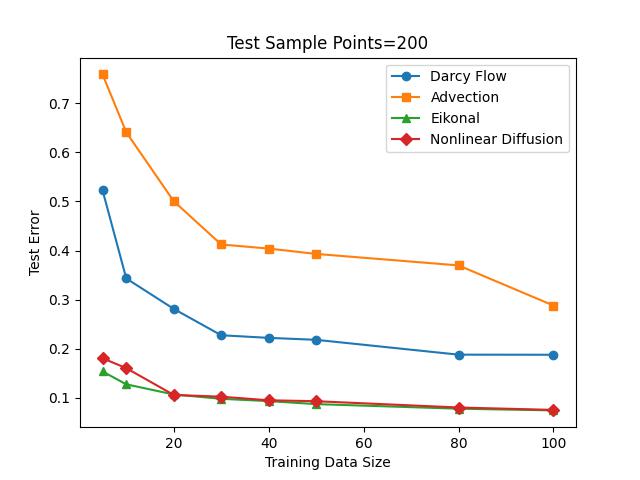}
    \end{subfigure}
    \caption{\small \textit{Test error trends across varying sample densities for PDE benchmarks}}
    \label{fig:Test-error-sample}
\end{figure}

\textbf{Data Efficiency.} From Table \ref{tab:test-trend-train} and Figure \ref{fig:test-error-train}, we use 4000 sample points and change different training data size to test the performance.
The error for Darcy Flow drops from 0.2005 to 0.0724, and the rapid improvement from 20 to 40 samples, then smaller incremental gains suggests the model captures the key inductive bias early.
The error for Eikonal decreases from 0.1112 to 0.0667, which is a moderate and smooth decline, indicating stable generalization.
In Figure \ref{fig:Test-error-sample}, it shows the test error for four PDE benchmarks as a function of training data size, under different test point resolutions.All PDEs generally show decreasing test error as the training data size increases, confirming that more training data improves generalization. Eikonal and Nonlinear Diffusion achieve good accuracy with very small training sets, suggesting these tasks require fewer data to generalize well. For Darcy Flow and Advection, the errors decrease more sharply with smaller training data sizes compared to larger ones which indicates that our model is more data-efficient in the data-scarce regime.

\begin{table}[H]
        \centering
    \caption{ \small \textit{Test errors under varying numbers of training data size with sample density=4000} }
    \small
    \label{tab:test-trend-train}
    \begin{tabular}{lccccc}
        \toprule
        \textit{Training data size} & 20     & 40     & 60     & 80     & 100    \\
        \midrule
        Darcy Flow        & 0.2005 & 0.1112 & 0.0905 & 0.0863 & 0.0724 \\
        Advection    & 0.4790 & 0.3381 & 0.3184 & 0.2380 & 0.1952 \\
        Eikonal          & 0.1112 & 0.0934 & 0.0762 & 0.0673 & 0.0667 \\
        Nonlinear Diffusion    & 0.0599 & 0.0400 & 0.0342 & 0.0312 & 0.0302 \\
        \bottomrule
    \end{tabular}
\end{table}

\textbf{Time Complexity and Memory Cost.} 
We analyze the computational complexity of the GOLA architecture in terms of the number of spatial points $N$, Fourier modes $M$, feature channels $C$, and edges $E \sim \mathcal{O}(Nk) $, where $k$ is the average number of neighbors in the sparse spatial graph.
The time complexity for GOLA is $\mathcal{O}(MNC)+\mathcal{O}(NkC^2)+\mathcal{O}(NkC)$.
The count of parameters for GOLA is 2,900,249.

\section{Conclusion}

We introduce a \textbf{G}raph-based \textbf{O}perator \textbf{L}earning with \textbf{A}ttention (GOLA) framework, that effectively models PDE solution operators on irregular domains using attention-enhanced graph neural networks and a Fourier-based encoder. Through comprehensive experiments across diverse PDE benchmarks including Darcy Flow, Advection, Eikonal, and Nonlinear Diffusion, GOLA consistently outperforms baselines particularly in data-scarce regimes and under severe resolution shifts. 
We demonstrate GOLA’s superior generalization, resolution scalability, and robustness to sparse sampling.
These results highlight the potential of combining spectral encoding and localized message passing with attention to build continuous, data-efficient operator approximators that adapt naturally to non-Euclidean geometries. This study demonstrates that graph-based representations provide a powerful and flexible foundation for advancing operator learning in real-world physical systems with irregular data.

\newpage

\input{GOLA.bbl}
\newpage
\appendix
\onecolumn

\newpage

\section{Additional Results}

\subsection{Test error trends across varying sample densities(30-100) for PDE benchmarks}

\begin{figure}[H]
    \centering

    \begin{subfigure}[b]{0.45\linewidth}
        \centering
        \includegraphics[width=\linewidth]{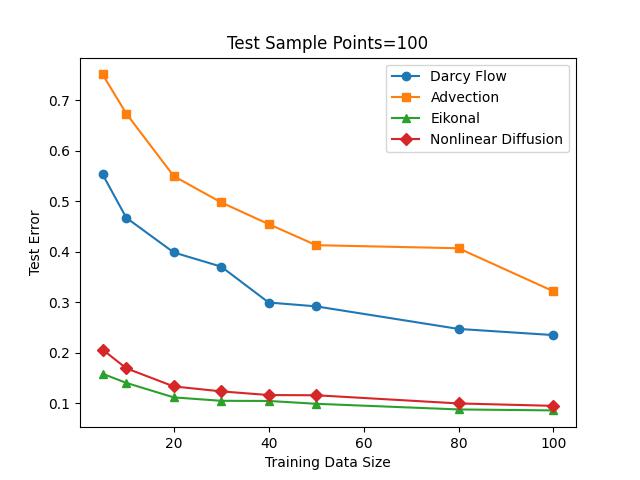}
    \end{subfigure}
    \hfill
    \begin{subfigure}[b]{0.45\linewidth}
        \centering
        \includegraphics[width=\linewidth]{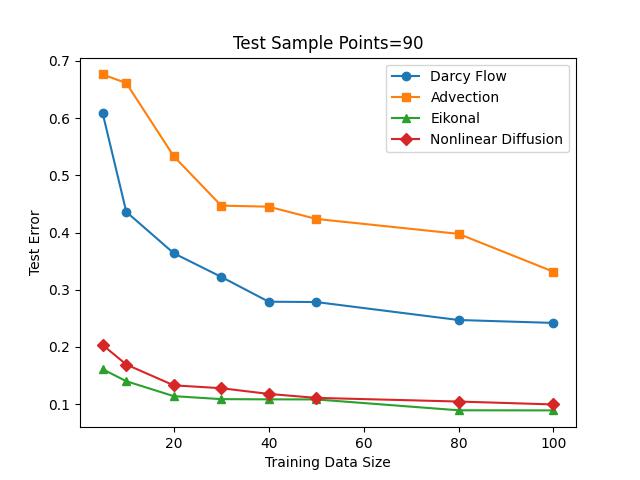}
    \end{subfigure}

    \begin{subfigure}[b]{0.45\linewidth}
        \centering
        \includegraphics[width=\linewidth]{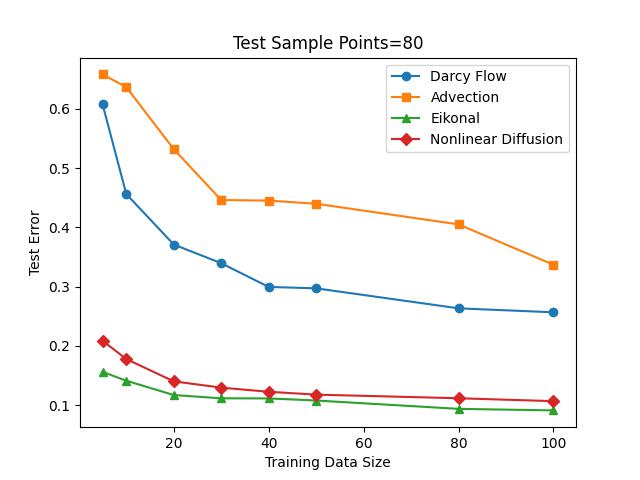}
    \end{subfigure}
    \hfill
    \begin{subfigure}[b]{0.45\linewidth}
        \centering
        \includegraphics[width=\linewidth]{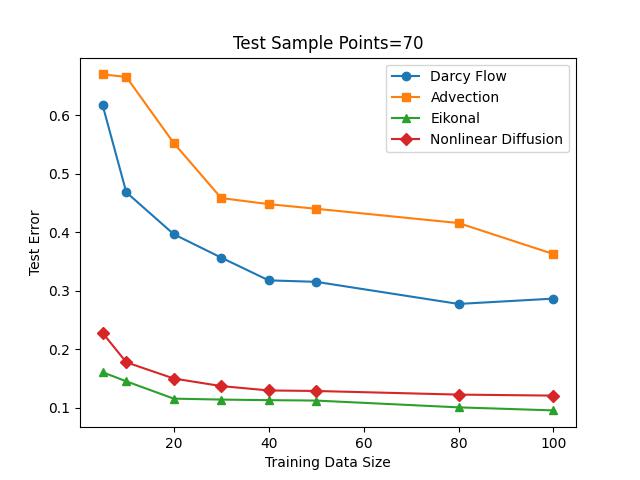}
    \end{subfigure}

    \begin{subfigure}[b]{0.45\linewidth}
        \centering
        \includegraphics[width=\linewidth]{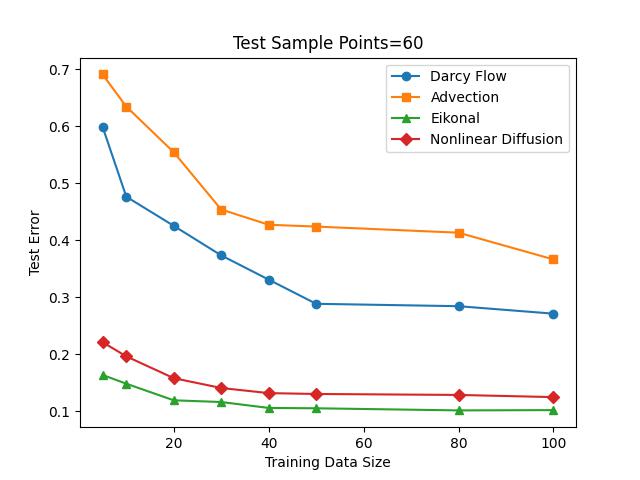}
    \end{subfigure}
    \hfill
    \begin{subfigure}[b]{0.45\linewidth}
        \centering
        \includegraphics[width=\linewidth]{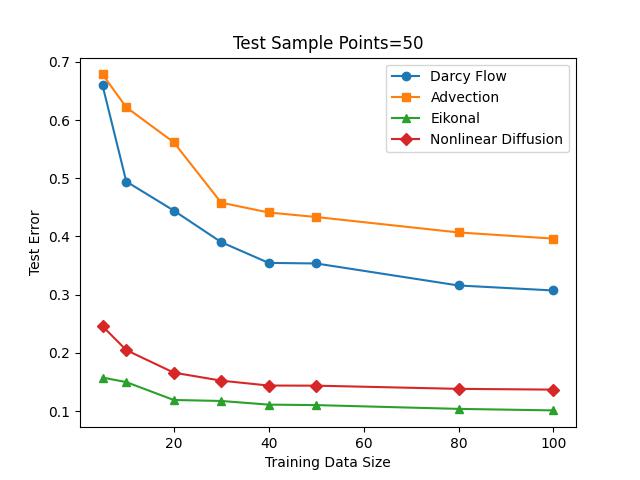}
    \end{subfigure}

    \begin{subfigure}[b]{0.45\linewidth}
        \centering
        \includegraphics[width=\linewidth]{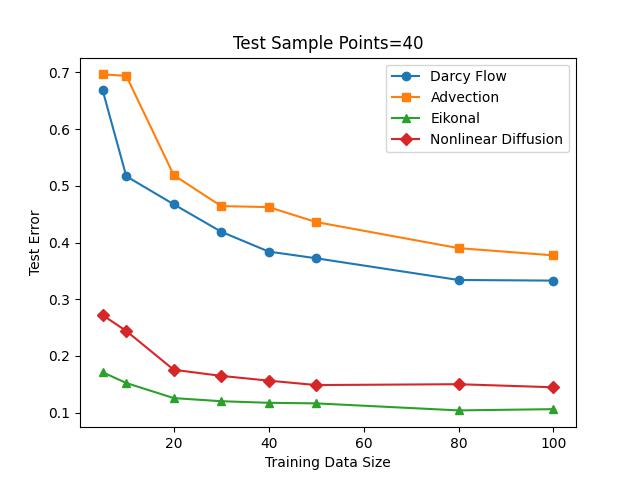}
    \end{subfigure}
    \hfill
    \begin{subfigure}[b]{0.45\linewidth}
        \centering
        \includegraphics[width=\linewidth]{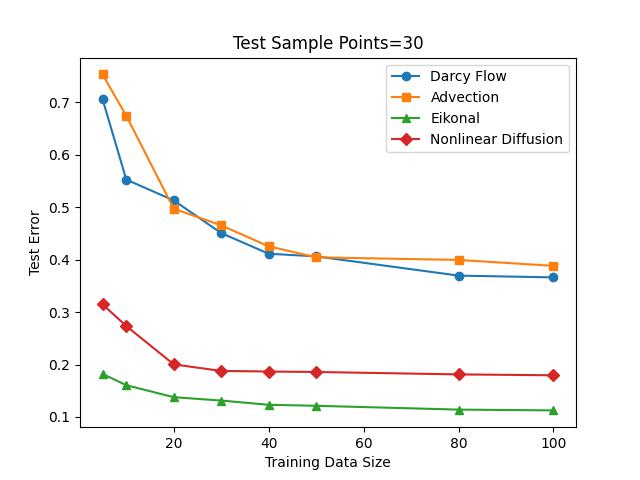}
    \end{subfigure}

    \caption{Test error trends across varying sample densities(30-100) for PDE benchmarks}
    \label{fig:Test-error-test100}
\end{figure}

In Figure \ref{fig:Test-error-test100}, it shows the test error for four PDE benchmarks as a function of training data size, under different test point resolutions ranging from 30 to 100.

\subsection{Error reduction on 30, 50, 100 training data across various sampling density}

From Table \ref{test-100}, \ref{test-50}, \ref{test-30}, with 100, 50, 30 training data size respectively, for each PDE benchmark, we choose sample density from 20 to 1000 to compare GKN and GOLA, and calculate the error reduction. It shows that our method is better than GKN and error reduction is significant.

\begin{table}[H]
\centering

\caption{Test errors trained on 100 training data size across various sampling density}

\end{table}

\begin{table}[H]
\centering
\caption{Test errors trained on 50 training data size across various sampling densities}
\label{test-50}
%
\end{table}

\begin{table}[H]
\centering
\caption{Test errors trained on 30 training data size across various sampling densities}
\label{test-30}
%
\end{table}

\subsection{Test errors across training data size for different sample density}
We choose sample density from 20 to 900, and for training data size from 5 to 100, we test on four PDE benchmarks as follows.

\begin{table*}

\begin{subtable}{\textwidth}
        \caption{\small \textit{Test sample density=900}}
        \centering

    \begin{tabular}{lcccccccc}
        \toprule
        \textit{Training data size} & 5      & 10     & 20     & 30     & 40     & 50    & 80   & 100    \\
        \midrule
        Darcy Flow      & 0.4996 & 0.2812 & 0.2215 & 0.1587 & 0.1503 & 0.1360 & 0.1230 & 0.1168\\
        Advection  & 0.6921 & 0.6862 & 0.4866 & 0.4074 & 0.3560 & 0.3460 & 0.2601 & 0.2310 \\
        Eikonal         & 0.1540 & 0.1245 & 0.1059 & 0.0760 & 0.0718 & 0.0704 & 0.0694 & 0.0655\\
        Nonlinear Diffusion  & 0.1589 & 0.1420 & 0.0740 & 0.0683 & 0.0595 & 0.0567 & 0.0474 & 0.0463\\
        \bottomrule
    \end{tabular}

\end{subtable}

\begin{subtable}{\textwidth}
        \caption{\small \textit{Test sample density=800}}
        \centering

  \begin{tabular}{lcccccccc}
    \toprule
    \textit{Training data size} & 5      & 10     & 20     & 30     & 40     & 50     & 80     & 100    \\
    \midrule
    Darcy Flow      & 0.5102 & 0.2781 & 0.2303 & 0.1576 & 0.1572 & 0.1381 & 0.1243 & 0.1226 \\
    Advection   & 0.7407 & 0.7109 & 0.4799 & 0.4148 & 0.3789 & 0.3641 & 0.2791 & 0.2408 \\
    Eikonal         & 0.1606 & 0.1210 & 0.1038 & 0.0756 & 0.0731 & 0.0730 & 0.0697 & 0.0625 \\
    Nonlinear Diffusion  & 0.1623 & 0.1406 & 0.0728 & 0.0706 & 0.0701 & 0.0652 & 0.0501 & 0.0469 \\
    \bottomrule
  \end{tabular}

\end{subtable}

\begin{subtable}{\textwidth}
        \caption{\small \textit{Test sample density=700}}
        \centering

  \begin{tabular}{lcccccccc}
    \toprule
    \textit{Training data size} & 5      & 10     & 20     & 30                           & 40     & 50                           & 80                           & 100    \\
    \midrule
    Darcy Flow           & 0.4985 & 0.2768 & 0.2332 & 0.1659 & 0.1592 & 0.1506 & 0.1301 & 0.1235 \\
    Advection            & 0.6978 & 0.6590 & 0.5594 & 0.4172 & 0.3747 & 0.3364 & 0.2926 & 0.2421 \\
    Eikonal              & 0.1605 & 0.1180 & 0.1038 & 0.0801 & 0.0755 & 0.0713 & 0.0643 & 0.0635 \\
    Nonlinear Diffusion  & 0.1642 & 0.1363 & 0.0731 & 0.0696 & 0.0623 & 0.0575 & 0.0548       & 0.0507 \\
    \bottomrule
  \end{tabular}

\end{subtable}

\begin{subtable}{\textwidth}
        \caption{\small \textit{Test sample density=600}}
        \centering

  \begin{tabular}{lcccccccc}
    \toprule
    \textit{Training data size}
      & 5      & 10     & 20     & 30     & 40     & 50  & 80   & 100    \\
    \midrule
    
    Darcy Flow          & 0.4817 & 0.2886 & 0.2362 & 0.1646 & 0.1616 & 0.1497 & 0.1401 & 0.1376 \\
    Advection           & 0.7537 & 0.7271 & 0.5754 & 0.4109 & 0.3661 & 0.3378 & 0.3216 & 0.2453 \\
    Eikonal             & 0.1483 & 0.1255 & 0.1073 & 0.0776 & 0.0772 & 0.0745 & 0.0691 & 0.0665 \\
    Nonlinear Diffusion & 0.2149 & 0.1440 & 0.0781 & 0.0709 & 0.0644 & 0.0618 & 0.0583 & 0.0553 \\
    \bottomrule
  \end{tabular}

\end{subtable}

\begin{subtable}{\textwidth}
        \caption{\small \textit{Test sample density=500}}
        \centering

\begin{tabular}{lcccccccc}
\toprule
\textit{Training data size} & 5     & 10    & 20    & 30    & 40    & 50    & 80 & 100   \\
\midrule
Darcy Flow             & 0.4855 & 0.2842 & 0.2247 & 0.1772 & 0.1643 & 0.1609 & 0.1414 & 0.1315 \\
Advection              & 0.7947 & 0.7011 & 0.5431 & 0.4089 & 0.3609 & 0.3468 & 0.3299 & 0.2462 \\
Eikonal                & 0.1537 & 0.1155 & 0.1060 & 0.0901 & 0.0811 & 0.0702 & 0.0689 & 0.0671 \\
Nonlinear Diffusion    & 0.1686 & 0.1464 & 0.0801 & 0.0773 & 0.0760 & 0.0733 & 0.0626 & 0.0594 \\
\bottomrule
\end{tabular}

\end{subtable}

\begin{subtable}{\textwidth}
        \caption{\small \textit{Test sample density=400}}
        \centering

\begin{tabular}{lcccccccc}
\toprule
\textit{Training data size} & 5      & 10     & 20     & 30     & 40     & 50  & 80   & 100    \\
\midrule
Darcy Flow           & 0.4970 & 0.3109 & 0.2471 & 0.1864 & 0.1804 & 0.1741 & 0.1481 & 0.1415 \\
Advection            & 0.7604 & 0.6733 & 0.5317 & 0.4090 & 0.4005 & 0.3658 & 0.3499 & 0.2516 \\
Eikonal              & 0.1547 & 0.1203 & 0.1058 & 0.0908 & 0.0883 & 0.0829 & 0.0725 & 0.0669 \\
Nonlinear Diffusion  & 0.1695 & 0.1411 & 0.0847 & 0.0815 & 0.0776 & 0.0766 & 0.0637 & 0.0618 \\
\bottomrule
\end{tabular}

\end{subtable}

\end{table*}

\begin{table*}

\begin{subtable}{\textwidth}
        \caption{\small \textit{Test sample density=300}}
        \centering

\begin{tabular}{lcccccccc}
\toprule
\textit{Training data size} & 5      & 10     & 20     & 30     & 40     & 50   & 80   & 100    \\
\midrule
Darcy Flow           & 0.5075 & 0.3233 & 0.2416 & 0.1906 & 0.1880 & 0.1771 & 0.1592 & 0.1479 \\
Advection            & 0.7606 & 0.6674 & 0.5323 & 0.4191 & 0.4178 & 0.3882 & 0.3505 & 0.2537 \\
Eikonal              & 0.1533 & 0.1246 & 0.1066 & 0.0921 & 0.0835 & 0.0824 & 0.0709 & 0.0664 \\
Nonlinear Diffusion  & 0.1722 & 0.1498 & 0.0920 & 0.0873 & 0.0849 & 0.0818 & 0.0694 & 0.0674 \\
\bottomrule
\end{tabular}

\end{subtable}

\begin{subtable}{\textwidth}
        \caption{\small \textit{Test sample density=200}}
        \centering
\begin{tabular}{lcccccccc}
\toprule
\textit{Training data size} & 5      & 10     & 20     & 30     & 40     & 50        &80              & 100    \\
\midrule
Darcy Flow           & 0.5227 & 0.3431 & 0.2812 & 0.2275 & 0.2222 & 0.2182                  & 0.1880 & 0.1878 \\
Advection            & 0.7586 & 0.6407 & 0.5004 & 0.4123 & 0.4040 & 0.3931      & 0.3697 & 0.2880 \\
Eikonal              & 0.1538 & 0.1276 & 0.1069 & 0.0981 & 0.0934 & 0.0873                  & 0.0780 & 0.0745 \\
Nonlinear Diffusion  & 0.1807 & 0.1605 & 0.1060 & 0.1024 & 0.0951 & 0.0932                  & 0.0804 & 0.0755 \\
\bottomrule
\end{tabular}

\end{subtable}

    \begin{subtable}{\textwidth}
        \caption{\small \textit{Test sample density=100}}
        \centering
\begin{tabular}{lcccccccc}
\toprule
\textit{Training data size} &5 & 10     & 20     & 30     & 40     & 50   &80  & 100    \\
\midrule
Darcy Flow           & 0.5534 & 0.4673 & 0.3986 & 0.3705 & 0.2994 & 0.2918 & 0.2471 & 0.2349 \\
Advection            & 0.7513 & 0.6738 & 0.5498 & 0.4977 & 0.4547 & 0.4130 & 0.4069 & 0.3216 \\
Eikonal              & 0.1583 & 0.1402 & 0.1115 & 0.1047 & 0.1044 & 0.0988 & 0.0876 & 0.0857 \\
Nonlinear Diffusion  & 0.2062 & 0.1691 & 0.1332 & 0.1234 & 0.1161 & 0.1157 & 0.0995 & 0.0947 \\
\bottomrule
\end{tabular}

\end{subtable}

    \begin{subtable}{\textwidth}
        \caption{\small \textit{Test sample density=90}}
        \centering
\begin{tabular}{lcccccccc}
\toprule
\textit{Training data size} & 5      & 10     & 20     & 30     & 40     & 50   & 80  & 100    \\
\midrule
Darcy Flow           & 0.6094 & 0.4362 & 0.3638 & 0.3225 & 0.2791 & 0.2785 & 0.2471 & 0.2420 \\
Advection            & 0.6764 & 0.6612 & 0.5334 & 0.4470 & 0.4452 & 0.4238 & 0.3977 & 0.3315 \\
Eikonal              & 0.1612 & 0.1401 & 0.1140 & 0.1088 & 0.1084 & 0.1083 & 0.0893 & 0.0891 \\
Nonlinear Diffusion  & 0.2036 & 0.1693 & 0.1329 & 0.1279 & 0.1179 & 0.1110 & 0.1046 & 0.0995 \\
\bottomrule
\end{tabular}

\end{subtable}

    \begin{subtable}{\textwidth}
        \caption{\small \textit{Test sample density=80}}
        \centering
\begin{tabular}{lcccccccc}
\toprule
\textit{Training data size} & 5      & 10     & 20     & 30     & 40     & 50   & 80  & 100    \\
\midrule
Darcy Flow           & 0.6076 & 0.4561 & 0.3709 & 0.3394 & 0.2994 & 0.2970 & 0.2633 & 0.2566 \\
Advection            & 0.6579 & 0.6365 & 0.5318 & 0.4460 & 0.4450 & 0.4397 & 0.4049 & 0.3368 \\
Eikonal              & 0.1561 & 0.1412 & 0.1170 & 0.1115 & 0.1113 & 0.1078 & 0.0937 & 0.0911 \\
Nonlinear Diffusion  & 0.2086 & 0.1777 & 0.1401 & 0.1295 & 0.1225 & 0.1177 & 0.1116 & 0.1067 \\
\bottomrule
\end{tabular}

\end{subtable}

    \begin{subtable}{\textwidth}
        \caption{\small \textit{Test sample density=70}}
        \centering
\begin{tabular}{lcccccccc}
\toprule
\textit{Training data size} & 5      & 10     & 20     & 30     & 40     & 50   & 80  & 100    \\
\midrule
Darcy Flow           & 0.6176 & 0.4682 & 0.3965 & 0.3565 & 0.3176 & 0.3151 & 0.2773 & 0.2864 \\
Advection            & 0.6697 & 0.6654 & 0.5524 & 0.4583 & 0.4479 & 0.4400 & 0.4156 & 0.3626 \\
Eikonal              & 0.1606 & 0.1450 & 0.1154 & 0.1138 & 0.1128 & 0.1121 & 0.1005 & 0.0953 \\
Nonlinear Diffusion  & 0.2279 & 0.1776 & 0.1498 & 0.1369 & 0.1294 & 0.1285 & 0.1223 & 0.1206 \\
\bottomrule
\end{tabular}

\end{subtable}

    \begin{subtable}{\textwidth}
        \caption{\small \textit{Test sample density=60}}
        \centering
\begin{tabular}{lcccccccc}
\toprule
\textit{Training data size} & 5      & 10     & 20     & 30     & 40                     & 50         & 80            & 100    \\
\midrule
Darcy Flow           & 0.5989 & 0.4762 & 0.4251 & 0.3733 & 0.3308                 & 0.2884                 & 0.2842 & 0.2711 \\
Advection            & 0.6906 & 0.6342 & 0.5541 & 0.4537 & 0.4270                 & 0.4238                 & 0.4131 & 0.3662 \\
Eikonal              & 0.1637 & 0.1483 & 0.1192 & 0.1162 & 0.1058   & 0.1052   & 0.1015 & 0.1020  \\
Nonlinear Diffusion  & 0.2211 & 0.1963 & 0.1580 & 0.1408 & 0.1318                 & 0.1303                 & 0.1286 & 0.1248 \\
\bottomrule
\end{tabular}

\end{subtable}

\end{table*}

\begin{table*}

    \begin{subtable}{\textwidth}
        \caption{\small \textit{Test sample density=50}}
        \centering
\begin{tabular}{lcccccccc}
\toprule
\textit{Training data size} & 5      & 10     & 20     & 30     & 40                     & 50     & 80                & 100    \\
\midrule
Darcy Flow           & 0.6595 & 0.4944 & 0.4445 & 0.3902 & 0.3546                 & 0.3536                 & 0.3158 & 0.3072 \\
Advection            & 0.6784 & 0.6219 & 0.5618 & 0.4580 & 0.4412                 & 0.4334                 & 0.4069 & 0.3747 \\
Eikonal              & 0.1576 & 0.1498 & 0.1191 & 0.1174 & 0.1111   & 0.1103   & 0.1038 & 0.0981 \\
Nonlinear Diffusion  & 0.2458 & 0.2048 & 0.1661 & 0.1523 & 0.1439                 & 0.1437                 & 0.1382 & 0.1369 \\
\bottomrule
\end{tabular}

\end{subtable}

    \begin{subtable}{\textwidth}
        \caption{\small \textit{Test sample density=40}}
        \centering
\begin{tabular}{lcccccccc}
\toprule
\textit{Training data size} & 5      & 10     & 20     & 30     & 40     & 50 & 80    & 100    \\
\midrule
Darcy Flow           & 0.6693 & 0.5169 & 0.4674 & 0.4190 & 0.3840 & 0.3722 &0.3339 & 0.3328 \\
Advection            & 0.6966 & 0.6941 & 0.5189 & 0.4642 & 0.4625 & 0.4361 &0.3901 & 0.3773 \\
Eikonal              & 0.1712 & 0.1521 & 0.1255 & 0.1200 & 0.1172 & 0.1162 &0.1038 & 0.1060 \\
Nonlinear Diffusion  & 0.2720 & 0.2439 & 0.1755 & 0.1647 & 0.1563 & 0.1485 &0.1500 & 0.1446 \\
\bottomrule
\end{tabular}

\end{subtable}

    \begin{subtable}{\textwidth}
        \caption{\small \textit{Test sample density=30}}
        \centering
\begin{tabular}{lcccccccc}
\toprule
\textit{Training data size} & 5      & 10     & 20     & 30     & 40     & 50  & 80   & 100    \\
\midrule
Darcy Flow           & 0.7064 & 0.5529 & 0.5131 & 0.4509 & 0.4114 & 0.4064 & 0.3697 & 0.3663 \\
Advection            & 0.7536 & 0.6746 & 0.4973 & 0.4655 & 0.4256 & 0.4046 & 0.3996 & 0.3883 \\
Eikonal              & 0.1822 & 0.1606 & 0.1377 & 0.1313 & 0.1232 & 0.1213 & 0.1140 & 0.1125 \\
Nonlinear Diffusion  & 0.3148 & 0.2735 & 0.2003 & 0.1878 & 0.1867 & 0.1858 & 0.1812 & 0.1795 \\
\bottomrule
\end{tabular}

\end{subtable}


       \begin{subtable}{\textwidth}
        \caption{\small \textit{Test sample density=20}}
        \centering
\begin{tabular}{lcccccccc}
\toprule
\textit{Training data size} &5 & 10     & 20     & 30     & 40     & 50  & 80   & 100    \\
\midrule
Darcy Flow           & 0.7416& 0.5963 & 0.5363 & 0.4984 & 0.4567 & 0.4484 & 0.4176 & 0.4073 \\
Advection            & 0.8587 & 0.7329 & 0.5114 & 0.4798 & 0.4661 & 0.4338 & 0.4186 & 0.3980 \\
Eikonal              & 0.1875 & 0.1797 & 0.1543 & 0.1402 & 0.1392 & 0.1361 & 0.1292 & 0.1236 \\
Nonlinear Diffusion  & 0.3202 & 0.3074 & 0.2403 & 0.2365 & 0.2297 & 0.2202 & 0.2000 & 0.1958 \\
\bottomrule
\end{tabular}

\end{subtable}

\end{table*}

\newpage

\subsection{Error reductions for training data size=40}

\begin{table*}[H]

        \begin{subtable}{\textwidth}
        \caption{\small \textit{Sample density=100, train data size=40}}
        \centering
    \begin{tabular}{lccc}
        \toprule
        \textit{Dataset}        & GKN     & Ours    & Error Reduction \\
        \midrule
        Darcy Flow              & 0.4122  & 0.2994  & 27.37\% \\
        Advection               & 0.4938  & 0.4547  & 7.92\%  \\
        Eikonal                 & 0.1668  & 0.1102  & 33.93\% \\
        Nonlinear Diffusion     & 0.1947  & 0.1161  & 40.37\% \\
        Poisson                 & 0.3754  & 0.3707  & 1.25\%  \\
        \bottomrule
    \end{tabular}

\end{subtable}

        \begin{subtable}{\textwidth}
        \caption{\small \textit{Sample density=200, train data size=40}}
        \centering
  \begin{tabular}{@{}lccc@{}}
    \toprule
    \textit{Dataset}         & GKN & Ours   & Error Reduction \\ 
    \midrule
    Darcy Flow               & 0.3281   & 0.2388 & 27.22\%         \\
    Advection                & 0.4840   & 0.4420 & 8.68\%          \\
    Eikonal                  & 0.1429   & 0.1050 & 26.52\%         \\
    Nonlinear Diffusion      & 0.1714   & 0.0951 & 44.52\%         \\
    \bottomrule
  \end{tabular}

\end{subtable}

        \begin{subtable}{\textwidth}
        \caption{\small \textit{Sample density=300, train data size=40}}
        \centering
  \begin{tabular}{@{}lccc@{}}
    \toprule
    \textit{Dataset}         & GKN & Ours   & Error Reduction \\ 
    \midrule
    Darcy Flow               & 0.2746   & 0.2126 & 22.58\%         \\
    Advection                & 0.4660   & 0.4178 & 10.34\%         \\
    Eikonal                  & 0.1390   & 0.0982 & 29.35\%         \\
    Nonlinear Diffusion      & 0.1589   & 0.0849 & 46.57\%         \\
    \bottomrule
  \end{tabular}

\end{subtable}

            \begin{subtable}{\textwidth}
        \caption{\small \textit{Sample density=400, train data size=40}}
        \centering
  \begin{tabular}{@{}lccc@{}}
    \toprule
    \textit{Dataset}         & GKN & Ours   & Error Reduction \\ 
    \midrule
    Darcy Flow               & 0.2446   & 0.1968 & 19.54\%         \\
    Advection                & 0.4590   & 0.4005 & 12.75\%         \\
    Eikonal                  & 0.1396   & 0.0952 & 31.81\%         \\
    Nonlinear Diffusion      & 0.1536   & 0.0776 & 49.48\%         \\
    \bottomrule
  \end{tabular}

\end{subtable}

            \begin{subtable}{\textwidth}
        \caption{\small \textit{Sample density=500, train data size=40}}
        \centering
  \begin{tabular}{@{}lccc@{}}
    \toprule
    \textit{Dataset}         & GKN & Ours   & Error Reduction \\ 
    \midrule
    Darcy Flow               & 0.2505   & 0.1878 & 25.03\%         \\
    Advection                & 0.4746   & 0.3609 & 23.96\%         \\
    Eikonal                  & 0.1355   & 0.0957 & 29.37\%         \\
    Nonlinear Diffusion      & 0.1389   & 0.0760 & 45.28\%         \\
    \bottomrule
  \end{tabular}

\end{subtable}

            \begin{subtable}{\textwidth}
        \caption{\small \textit{Sample density=600, train data size=40}}
        \centering
  \begin{tabular}{@{}lccc@{}}
    \toprule
    \textit{Dataset}         & GKN & Ours   & Error Reduction \\ 
    \midrule
    Darcy Flow               & 0.2416   & 0.1648 & 31.79\%         \\
    Advection                & 0.4352   & 0.3661 & 15.88\%         \\
    Eikonal                  & 0.1323   & 0.0772 & 41.65\%         \\
    Nonlinear Diffusion      & 0.1441   & 0.0739 & 48.72\%         \\
    \bottomrule
  \end{tabular}

\end{subtable}

            \begin{subtable}{\textwidth}
        \caption{\small \textit{Sample density=700, train data size=40}}
        \centering
  \begin{tabular}{@{}lccc@{}}
    \toprule
    \textit{Dataset}         & GKN & Ours   & Error Reduction \\ 
    \midrule
    Darcy Flow               & 0.2257   & 0.1592 & 29.46\%         \\
    Advection                & 0.4546   & 0.3747 & 17.58\%         \\
    Eikonal                  & 0.1294   & 0.0819 & 36.71\%         \\
    Nonlinear Diffusion      & 0.1424   & 0.0731 & 48.67\%         \\
    \bottomrule
  \end{tabular}

\end{subtable}

\end{table*}

\begin{table*}

            \begin{subtable}{\textwidth}
        \caption{\small \textit{Sample density=800, train data size=40}}
        \centering
  \begin{tabular}{@{}lccc@{}}
    \toprule
    \textit{Dataset}         & GKN & Ours   & Error Reduction \\ 
    \midrule
    Darcy Flow               & 0.2120   & 0.1572 & 25.85\%         \\
    Advection                & 0.4558   & 0.3789 & 16.87\%         \\
    Eikonal                  & 0.1304   & 0.0761 & 41.64\%         \\
    Nonlinear Diffusion      & 0.1413   & 0.0701 & 50.39\%         \\
    \bottomrule
  \end{tabular}

\end{subtable}

            \begin{subtable}{\textwidth}
        \caption{\small \textit{Sample density=900, train data size=40}}
        \centering
  \begin{tabular}{@{}lccc@{}}
    \toprule
    \textit{Dataset}         & GKN & Ours   & Error Reduction \\ 
    \midrule
    Darcy Flow               & 0.2068   & 0.1503 & 27.32\%         \\
    Advection                & 0.4292   & 0.3560 & 17.05\%         \\
    Eikonal                  & 0.1306   & 0.0805 & 38.36\%         \\
    Nonlinear Diffusion      & 0.1441   & 0.0691 & 52.05\%         \\
    \bottomrule
  \end{tabular}

\end{subtable}

            \begin{subtable}{\textwidth}
        \caption{\small \textit{Sample density=1000, train data size=40}}
        \centering
  \begin{tabular}{@{}lccc@{}}
    \toprule
    \textit{Dataset}         & GKN & Ours   & Error Reduction \\ 
    \midrule
    Darcy Flow               & 0.2257   & 0.1592 & 29.46\%         \\
    Advection                & 0.4546   & 0.3747 & 17.58\%         \\
    Eikonal                  & 0.1294   & 0.0819 & 36.71\%         \\
    Nonlinear Diffusion      & 0.1424   & 0.0731 & 48.67\%         \\
    \bottomrule
  \end{tabular}

\end{subtable}

            \begin{subtable}{\textwidth}
        \caption{\small \textit{Sample density=20, train data size=40}}
        \centering
  \begin{tabular}{@{}lccc@{}}
    \toprule
    \textit{Dataset}         & GKN & Ours   & Error Reduction \\ 
    \midrule
    Darcy Flow               & 0.5527   & 0.4567 & 17.37\%         \\
    Advection                & 0.9735   & 0.5219 & 46.39\%         \\
    Eikonal                  & 0.1868   & 0.1451 & 22.32\%         \\
    Nonlinear Diffusion      & 0.2892   & 0.2297 & 20.57\%         \\
    \bottomrule
  \end{tabular}

\end{subtable}

              \begin{subtable}{\textwidth}
        \caption{\small \textit{Sample density=30, train data size=40}}
        \centering
  \begin{tabular}{@{}lccc@{}}
    \toprule
    \textit{Dataset}         & GKN & Ours   & Error Reduction \\ 
    \midrule
    Darcy Flow               & 0.5180   & 0.4114 & 20.58\%         \\
    Advection                & 0.7502   & 0.4694 & 37.43\%         \\
    Eikonal                  & 0.1847   & 0.1354 & 26.69\%         \\
    Nonlinear Diffusion      & 0.2638   & 0.1867 & 29.23\%         \\
    \bottomrule
  \end{tabular}

\end{subtable}

              \begin{subtable}{\textwidth}
        \caption{\small \textit{Sample density=40, train data size=40}}
        \centering
  \begin{tabular}{@{}lccc@{}}
    \toprule
    \textit{Dataset}         & GKN & Ours   & Error Reduction \\ 
    \midrule
    Darcy Flow               & 0.5074   & 0.3840 & 24.32\%         \\
    Advection                & 0.6341   & 0.4714 & 25.66\%         \\
    Eikonal                  & 0.1833   & 0.1234 & 32.68\%         \\
    Nonlinear Diffusion      & 0.2464   & 0.1563 & 36.57\%         \\
    \bottomrule
  \end{tabular}

\end{subtable}

              \begin{subtable}{\textwidth}
        \caption{\small \textit{Sample density=50, train data size=40}}
        \centering
  \begin{tabular}{@{}lccc@{}}
    \toprule
    \textit{Dataset}         & GKN & Ours   & Error Reduction \\ 
    \midrule
    Darcy Flow               & 0.4819   & 0.3546 & 26.42\%         \\
    Advection                & 0.5321   & 0.4897 & 7.97\%          \\
    Eikonal                  & 0.1798   & 0.1194 & 33.59\%         \\
    Nonlinear Diffusion      & 0.2417   & 0.1439 & 40.46\%         \\
    \bottomrule
  \end{tabular}

\end{subtable}

\end{table*}

\begin{table*}

              \begin{subtable}{\textwidth}
        \caption{\small \textit{Sample density=60, train data size=40}}
        \centering
  \begin{tabular}{@{}lccc@{}}
    \toprule
    \textit{Dataset}         & GKN & Ours   & Error Reduction \\ 
    \midrule
    Darcy Flow               & 0.4664   & 0.3308 & 29.07\%         \\
    Advection                & 0.5688   & 0.4993 & 12.22\%         \\
    Eikonal                  & 0.1777   & 0.1181 & 33.54\%         \\
    Nonlinear Diffusion      & 0.2349   & 0.1318 & 43.89\%         \\
    \bottomrule
  \end{tabular}

\end{subtable}

              \begin{subtable}{\textwidth}
        \caption{\small \textit{Sample density=70, train data size=40}}
        \centering
  \begin{tabular}{@{}lccc@{}}
    \toprule
    \textit{Dataset}         & GKN      & Ours   & Error Reduction \\ 
    \midrule
    Darcy Flow               & 0.4518   & 0.3176 & 29.70\%         \\
    Advection                & 0.5410   & 0.4766 & 11.90\%         \\
    Eikonal                  & 0.1663   & 0.1147 & 31.03\%         \\
    Nonlinear Diffusion      & 0.2195   & 0.1294 & 41.05\%         \\
    \bottomrule
  \end{tabular}

\end{subtable}

              \begin{subtable}{\textwidth}
        \caption{\small \textit{Sample density=80, train data size=40}}
        \centering
  \begin{tabular}{@{}lccc@{}}
    \toprule
    \textit{Dataset}         & GKN      & Ours   & Error Reduction \\ 
    \midrule
    Darcy Flow               & 0.4558   & 0.2994 & 34.31\%         \\
    Advection                & 0.5175   & 0.4535 & 12.37\%         \\
    Eikonal                  & 0.1642   & 0.1132 & 31.06\%         \\
    Nonlinear Diffusion      & 0.2127   & 0.1225 & 42.41\%         \\
    \bottomrule
  \end{tabular}

\end{subtable}

              \begin{subtable}{\textwidth}
        \caption{\small \textit{Sample density=90, train data size=40}}
        \centering
  \begin{tabular}{@{}lccc@{}}
    \toprule
    \textit{Dataset}         & GKN      & Ours   & Error Reduction \\ 
    \midrule
    Darcy Flow               & 0.4378   & 0.2791 & 36.25\%         \\
    Advection                & 0.5039   & 0.4452 & 11.65\%         \\
    Eikonal                  & 0.1706   & 0.1122 & 34.23\%         \\
    Nonlinear Diffusion      & 0.1996   & 0.1179 & 40.93\%         \\
    \bottomrule
  \end{tabular}

\end{subtable}

\end{table*}

\newpage

\subsection{Error reductions for training data size=20 }

\begin{table*}

    \begin{subtable}{\textwidth}
        \caption{\small \textit{Sample density=100, train data size=20}}
        \centering
    \label{tab:Ntr20_Nte100_pts100}
    \begin{tabular}{lccc}
        \toprule
        \textit{Dataset}        & GKN     & Ours    & Error Reduction \\
        \midrule
        Darcy Flow              & 0.4790  & 0.3986  & 16.78\% \\
        Advection               & 0.8125  & 0.5498  & 32.33\% \\
        Eikonal                 & 0.1848  & 0.1115  & 39.66\% \\
        Nonlinear Diffusion     & 0.2284  & 0.1332  & 41.68\% \\
        \bottomrule
    \end{tabular}
    
\end{subtable}  

    \begin{subtable}{\textwidth}
        \caption{\small \textit{Sample density=200, train data size=20}}
        \centering
    \label{tab:Ntr20_Nte100_pts200}
    \begin{tabular}{lccc}
        \toprule
        \textit{Dataset}      & GKN      & Ours    & Error Reduction \\
        \midrule
        Darcy Flow            & 0.3937   & 0.2812  & 28.58\% \\
        Advection             & 0.8552   & 0.5004  & 41.49\% \\
        Eikonal               & 0.1780   & 0.1069  & 39.94\% \\
        Nonlinear Diffusion   & 0.2056   & 0.1060  & 48.44\% \\
        \bottomrule
    \end{tabular}

\end{subtable}  

    \begin{subtable}{\textwidth}
        \caption{\small \textit{Sample density=300, train data size=20}}
        \centering
    \label{tab:Ntr20_Nte100_pts300}
    \begin{tabular}{lccc}
        \toprule
        \textit{Dataset}      & GKN      & Ours    & Error Reduction \\
        \midrule
        Darcy Flow            & 0.3414   & 0.2416  & 29.23\% \\
        Advection             & 0.7282   & 0.5323  & 26.90\% \\
        Eikonal               & 0.1757   & 0.1066  & 39.33\% \\
        Nonlinear Diffusion   & 0.1923   & 0.0920  & 52.16\% \\
        \bottomrule
    \end{tabular}

\end{subtable}

    \begin{subtable}{\textwidth}
        \caption{\small \textit{Sample density=400, train data size=20}}
        \centering
    \label{tab:Ntr20_Nte100_pts400}
    \begin{tabular}{lccc}
        \toprule
        \textit{Dataset}      & GKN      & Ours    & Error Reduction \\
        \midrule
        Darcy Flow            & 0.3291   & 0.2471  & 24.92\% \\
        Advection             & 0.7667   & 0.5317  & 30.65\% \\
        Eikonal               & 0.1716   & 0.1058  & 38.34\% \\
        Nonlinear Diffusion   & 0.1831   & 0.0847  & 53.74\% \\
        \bottomrule
    \end{tabular}
    
\end{subtable}

    \begin{subtable}{\textwidth}
        \caption{\small \textit{Sample density=500, train data size=20}}
        \centering
    \label{tab:Ntr20_Nte100_pts500}
    \begin{tabular}{lccc}
        \toprule
        \textit{Dataset}      & GKN      & Ours    & Error Reduction \\
        \midrule
        Darcy Flow            & 0.3285   & 0.2247  & 31.60\% \\
        Advection             & 1.0101   & 0.5431  & 46.23\% \\
        Eikonal               & 0.1703   & 0.1060  & 37.76\% \\
        Nonlinear Diffusion   & 0.1783   & 0.0801  & 55.08\% \\
        \bottomrule
    \end{tabular}
    
\end{subtable} 

    \begin{subtable}{\textwidth}
        \caption{\small \textit{Sample density=600, train data size=20}}
        \centering
    \label{tab:Ntr20_Nte100_pts600}
    \begin{tabular}{lccc}
        \toprule
        \textit{Dataset}        & GKN     & Ours    & Error Reduction \\
        \midrule
        Darcy Flow              & 0.3030  & 0.2362  & 22.05\% \\
        Advection               & 0.7948  & 0.5754  & 27.60\% \\
        Eikonal                 & 0.1739  & 0.1073  & 38.30\% \\
        Nonlinear Diffusion     & 0.1780  & 0.0781  & 56.12\% \\
        \bottomrule
    \end{tabular}
    
\end{subtable} 

    \begin{subtable}{\textwidth}
        \caption{\small \textit{Sample density=700, train data size=20}}
        \centering
    \label{tab:Ntr20_Nte100_pts700}
    \begin{tabular}{lccc}
        \toprule
        \textit{Dataset}        & GKN      & Ours    & Error Reduction \\
        \midrule
        Darcy Flow              & 0.3001   & 0.2332  & 22.29\% \\
        Advection               & 0.7928   & 0.5594  & 29.44\% \\
        Eikonal                 & 0.1682   & 0.1038  & 38.29\% \\
        Nonlinear Diffusion     & 0.1755   & 0.0731  & 58.35\% \\
        \bottomrule
    \end{tabular}
    
\end{subtable} 

\end{table*}

\begin{table*}

    \begin{subtable}{\textwidth}
        \caption{\small \textit{Sample density=800, train data size=20}}
        \centering
    \label{tab:Ntr20_Nte100_pts800}
    \begin{tabular}{lccc}
        \toprule
        \textit{Dataset}        & GKN      & Ours    & Error Reduction \\
        \midrule
        Darcy Flow              & 0.2841   & 0.2303  & 18.94\% \\
        Advection               & 0.8037   & 0.4799  & 40.29\% \\
        Eikonal                 & 0.1696   & 0.1038  & 38.80\% \\
        Nonlinear Diffusion     & 0.1749   & 0.0728  & 58.38\% \\
        \bottomrule
    \end{tabular}
    
\end{subtable}

    \begin{subtable}{\textwidth}
        \caption{\small \textit{Sample density=900, train data size=20}}
        \centering
    \label{tab:Ntr20_Nte100_pts900}
    \begin{tabular}{lccc}
        \toprule
        \textit{Dataset}        & GKN      & Ours    & Error Reduction \\
        \midrule
        Darcy Flow              & 0.2787   & 0.2215  & 20.52\% \\
        Advection               & 0.7267   & 0.4866  & 33.04\% \\
        Eikonal                 & 0.1705   & 0.1059  & 37.89\% \\
        Nonlinear Diffusion     & 0.1755   & 0.0740  & 57.83\% \\
        \bottomrule
    \end{tabular}
    
\end{subtable} 

    \begin{subtable}{\textwidth}
        \caption{\small \textit{Sample density=1000, train data size=20}}
        \centering
    \label{tab:Ntr20_Nte100_pts1000}
    \begin{tabular}{lccc}
        \toprule
        \textit{Dataset}      & GKN      & Ours     & Error Reduction \\
        \midrule
        Darcy Flow            & 0.2660   & 0.2297   & 13.65\% \\
        Advection             & 0.7421   & 0.5268   & 29.01\% \\
        Eikonal               & 0.1697   & 0.1059   & 37.60\% \\
        Nonlinear Diffusion   & 0.1716   & 0.0757   & 55.89\% \\
        \bottomrule
    \end{tabular}
    
\end{subtable}

    \begin{subtable}{\textwidth}
        \caption{\small \textit{Sample density=20, train data size=20}}
        \centering
    \label{tab:Ntr20_Nte100_pts20}
    \begin{tabular}{lccc}
        \toprule
        \textit{Dataset}        & GKN      & Ours    & Error Reduction \\
        \midrule
        Darcy Flow              & 0.5909   & 0.5363  & 9.24\%  \\
        Advection               & 1.0704   & 0.5114  & 52.22\% \\
        Eikonal                 & 0.1985   & 0.1543  & 22.27\% \\
        Nonlinear Diffusion     & 0.3053   & 0.2403  & 21.29\% \\
        \bottomrule
    \end{tabular}
    
\end{subtable}

    \begin{subtable}{\textwidth}
        \caption{\small \textit{Sample density=30, train data size=20}}
        \centering
    \label{tab:Ntr20_Nte100_pts30}
    \begin{tabular}{lccc}
        \toprule
        \textit{Dataset}        & GKN      & Ours    & Error Reduction \\
        \midrule
        Darcy Flow              & 0.5712   & 0.5131  & 10.17\% \\
        Advection               & 1.0391   & 0.4973  & 52.14\% \\
        Eikonal                 & 0.1941   & 0.1377  & 29.06\% \\
        Nonlinear Diffusion     & 0.2804   & 0.2003  & 28.57\% \\
        \bottomrule
    \end{tabular}
    
\end{subtable}

    \begin{subtable}{\textwidth}
        \caption{\small \textit{Sample density=40, train data size=20}}
        \centering
    \label{tab:Ntr20_Nte100_pts40}
    \begin{tabular}{lccc}
        \toprule
        \textit{Dataset}      & GKN      & Ours    & Error Reduction \\
        \midrule
        Darcy Flow            & 0.5493   & 0.4674  & 14.91\% \\
        Advection             & 0.9612   & 0.5189  & 46.02\% \\
        Eikonal               & 0.1916   & 0.1255  & 34.50\% \\
        Nonlinear Diffusion   & 0.2655   & 0.1755  & 33.90\% \\
        \bottomrule
    \end{tabular}
    
\end{subtable} 

    \begin{subtable}{\textwidth}
        \caption{\small \textit{Sample density=50, train data size=20}}
        \centering
    \label{tab:Ntr20_Nte100_pts50}
    \begin{tabular}{lccc}
        \toprule
        \textit{Dataset}        & GKN      & Ours    & Error Reduction \\
        \midrule
        Darcy Flow              & 0.5221   & 0.4445  & 14.86\% \\
        Advection               & 0.9652   & 0.5618  & 41.79\% \\
        Eikonal                 & 0.1880   & 0.1191  & 36.65\% \\
        Nonlinear Diffusion     & 0.2525   & 0.1661  & 34.22\% \\
        \bottomrule
    \end{tabular}
    
\end{subtable} 

\end{table*}

\begin{table*}

    \begin{subtable}{\textwidth}
        \caption{\small \textit{Sample density=60, train data size=20}}
        \centering
    \label{tab:Ntr20_Nte100_pts60}
    \begin{tabular}{lccc}
        \toprule
        \textit{Dataset}        & GKN      & Ours    & Error Reduction \\
        \midrule
        Darcy Flow              & 0.5013   & 0.4251  & 15.20\% \\
        Advection               & 0.8556   & 0.5541  & 35.24\% \\
        Eikonal                 & 0.1864   & 0.1192  & 36.05\% \\
        Nonlinear Diffusion     & 0.2459   & 0.1580  & 35.75\% \\
        \bottomrule
    \end{tabular}
    
\end{subtable}

    \begin{subtable}{\textwidth}
        \caption{\small \textit{Sample density=70, train data size=20}}
        \centering
    \label{tab:Ntr20_Nte100_pts70}
    \begin{tabular}{lccc}
        \toprule
        \textit{Dataset}      & GKN      & Ours    & Error Reduction \\
        \midrule
        Darcy Flow            & 0.4975   & 0.3965  & 20.30\% \\
        Advection             & 0.8128   & 0.5524  & 32.04\% \\
        Eikonal               & 0.1838   & 0.1154  & 37.21\% \\
        Nonlinear Diffusion   & 0.2420   & 0.1498  & 38.10\% \\
        \bottomrule
    \end{tabular}
    
\end{subtable}

    \begin{subtable}{\textwidth}
        \caption{\small \textit{Sample density=80, train data size=20}}
        \centering
    \label{tab:Ntr20_Nte100_pts80}
    \begin{tabular}{lccc}
        \toprule
        \textit{Dataset}        & GKN      & Ours    & Error Reduction \\
        \midrule
        Darcy Flow              & 0.4937   & 0.3709  & 24.87\% \\
        Advection               & 0.7770   & 0.5318  & 31.56\% \\
        Eikonal                 & 0.1835   & 0.1170  & 36.24\% \\
        Nonlinear Diffusion     & 0.2387   & 0.1401  & 41.31\% \\
        \bottomrule
    \end{tabular}
    
\end{subtable} 

    \begin{subtable}{\textwidth}
        \caption{\small \textit{Sample density=90, train data size=20}}
        \centering
    \label{tab:Ntr20_Nte100_pts90}
    \begin{tabular}{lccc}
        \toprule
        \textit{Dataset}      & GKN      & Ours    & Error Reduction \\
        \midrule
        Darcy Flow            & 0.4920   & 0.3638  & 26.06\% \\
        Advection             & 0.8157   & 0.5334  & 34.61\% \\
        Eikonal               & 0.1847   & 0.1140  & 38.28\% \\
        Nonlinear Diffusion   & 0.2313   & 0.1329  & 42.54\% \\
        \bottomrule
    \end{tabular}
    
\end{subtable}

\end{table*}

\newpage

\subsection{Error reductions for training data size=10}

\begin{table*}

    \begin{subtable}{\textwidth}
        \caption{\small \textit{Sample density=100, train data size=10}}
        \centering
    \begin{tabular}{lccc}
        \toprule
        \textit{Dataset}      & GKN      & Ours    & Error Reduction \\
        \midrule
        Darcy Flow            & 0.5692   & 0.4673  & 17.90\% \\
        Advection             & 1.0858   & 0.6738  & 37.94\% \\
        Eikonal               & 0.2061   & 0.1402  & 31.98\% \\
        Nonlinear Diffusion   & 0.2531   & 0.1691  & 33.19\% \\
        \bottomrule
    \end{tabular}
    
\end{subtable}

    \begin{subtable}{\textwidth}
        \caption{\small \textit{Sample density=200, train data size=10}}
        \centering
    \begin{tabular}{lccc}
        \toprule
        \textit{Dataset}      & GKN      & Ours    & Error Reduction \\
        \midrule
        Darcy Flow            & 0.4593   & 0.3431  & 25.30\% \\
        Advection             & 1.0516   & 0.6407  & 39.07\% \\
        Eikonal               & 0.2018   & 0.1276  & 36.77\% \\
        Nonlinear Diffusion   & 0.2500   & 0.1605  & 35.80\% \\
        \bottomrule
    \end{tabular}
    
\end{subtable}

    \begin{subtable}{\textwidth}
        \caption{\small \textit{Sample density=300, train data size=10}}
        \centering
    \begin{tabular}{lccc}
        \toprule
        \textit{Dataset}      & GKN      & Ours    & Error Reduction \\
        \midrule
        Darcy Flow            & 0.4014   & 0.3233  & 19.46\% \\
        Advection             & 0.9618   & 0.6674  & 30.61\% \\
        Eikonal               & 0.1955   & 0.1246  & 36.27\% \\
        Nonlinear Diffusion   & 0.2548   & 0.1498  & 41.21\% \\
        \bottomrule
    \end{tabular}
    
\end{subtable}

    \begin{subtable}{\textwidth}
        \caption{\small \textit{Sample density=400, train data size=10}}
        \centering
    \begin{tabular}{lccc}
        \toprule
        \textit{Dataset}      & GKN      & Ours    & Error Reduction \\
        \midrule
        Darcy Flow            & 0.3750   & 0.3109  & 17.09\% \\
        Advection             & 1.0257   & 0.6733  & 34.36\% \\
        Eikonal               & 0.1913   & 0.1203  & 37.11\% \\
        Nonlinear Diffusion   & 0.2430   & 0.1411  & 41.93\% \\
        \bottomrule
    \end{tabular}
    
\end{subtable}

    \begin{subtable}{\textwidth}
        \caption{\small \textit{Sample density=500, train data size=10}}
        \centering
    \begin{tabular}{lccc}
        \toprule
        \textit{Dataset}      & GKN      & Ours    & Error Reduction \\
        \midrule
        Darcy Flow            & 0.3764   & 0.2842  & 24.50\% \\
        Advection             & 0.9785   & 0.7011  & 28.35\% \\
        Eikonal               & 0.1905   & 0.1155  & 39.37\% \\
        Nonlinear Diffusion   & 0.2383   & 0.1464  & 38.56\% \\
        \bottomrule
    \end{tabular}
    
\end{subtable}

    \begin{subtable}{\textwidth}
        \caption{\small \textit{Sample density=600, train data size=10}}
        \centering
    \begin{tabular}{lccc}
        \toprule
        \textit{Dataset}      & GKN      & Ours    & Error Reduction \\
        \midrule
        Darcy Flow            & 0.3645   & 0.2886  & 20.82\% \\
        Advection             & 1.0162   & 0.7271  & 28.45\% \\
        Eikonal               & 0.1929   & 0.1255  & 34.94\% \\
        Nonlinear Diffusion   & 0.2360   & 0.1440  & 38.98\% \\
        \bottomrule
    \end{tabular}
    
\end{subtable} 

    \begin{subtable}{\textwidth}
        \caption{\small \textit{Sample density=700, train data size=10}}
        \centering
    \begin{tabular}{lccc}
        \toprule
        \textit{Dataset}      & GKN      & Ours    & Error Reduction \\
        \midrule
        Darcy Flow            & 0.3522   & 0.2768  & 21.41\% \\
        Advection             & 0.9918   & 0.6590  & 33.56\% \\
        Eikonal               & 0.1929   & 0.1180  & 38.83\% \\
        Nonlinear Diffusion   & 0.2289   & 0.1363  & 40.45\% \\
        \bottomrule
    \end{tabular}
    
\end{subtable}

\end{table*}

\begin{table*}

    \begin{subtable}{\textwidth}
        \caption{\small \textit{Sample density=800, train data size=10}}
        \centering
    \begin{tabular}{lccc}
        \toprule
        \textit{Dataset}      & GKN      & Ours    & Error Reduction \\
        \midrule
        Darcy Flow            & 0.3400   & 0.2781  & 18.21\% \\
        Advection             & 1.0006   & 0.7109  & 28.95\% \\
        Eikonal               & 0.1909   & 0.1210  & 36.62\% \\
        Nonlinear Diffusion   & 0.2304   & 0.1406  & 38.98\% \\
        \bottomrule
    \end{tabular}
    
\end{subtable}

    \begin{subtable}{\textwidth}
        \caption{\small \textit{Sample density=900, train data size=10}}
        \centering
    \begin{tabular}{lccc}
        \toprule
        \textit{Dataset}        & GKN      & Ours    & Error Reduction \\
        \midrule
        Darcy Flow              & 0.3356   & 0.2812  & 16.21\% \\
        Advection               & 1.0374   & 0.6862  & 33.85\% \\
        Eikonal                 & 0.1913   & 0.1245  & 34.92\% \\
        Nonlinear Diffusion     & 0.2319   & 0.1420  & 38.77\% \\
        \bottomrule
    \end{tabular}
    
\end{subtable}

    \begin{subtable}{\textwidth}
        \caption{\small \textit{Sample density=1000, train data size=10}}
        \centering
    \begin{tabular}{lccc}
        \toprule
        \textit{Dataset}      & GKN      & Ours    & Error Reduction \\
        \midrule
        Darcy Flow            & 0.3345   & 0.2787  & 16.68\% \\
        Advection             & 1.0026   & 0.6668  & 33.49\% \\
        Eikonal               & 0.1956   & 0.1286  & 34.25\% \\
        Nonlinear Diffusion   & 0.2346   & 0.1204  & 48.68\% \\
        \bottomrule
    \end{tabular}
    
\end{subtable} 

    \begin{subtable}{\textwidth}
        \caption{\small \textit{Sample density=20, train data size=10}}
        \centering
    \begin{tabular}{lccc}
        \toprule
        \textit{Dataset}      & GKN      & Ours    & Error Reduction \\
        \midrule
        Darcy Flow            & 0.6578   & 0.5963  & 9.35\% \\
        Advection             & 1.1149   & 0.7329  & 34.26\% \\
        Eikonal               & 0.2144   & 0.1797  & 16.18\% \\
        Nonlinear Diffusion   & 0.3280   & 0.3074  & 6.28\% \\
        \bottomrule
    \end{tabular}
    
\end{subtable}

    \begin{subtable}{\textwidth}
        \caption{\small \textit{Sample density=30, train data size=10}}
        \centering
    \begin{tabular}{lccc}
        \toprule
        \textit{Dataset}      & GKN      & Ours    & Error Reduction \\
        \midrule
        Darcy Flow            & 0.6580   & 0.5529  & 15.97\% \\
        Advection             & 1.1036   & 0.6746  & 38.87\% \\
        Eikonal               & 0.2073   & 0.1606  & 22.53\% \\
        Nonlinear Diffusion   & 0.3017   & 0.2735  & 9.35\%  \\
        \bottomrule
    \end{tabular}
    
\end{subtable} 

    \begin{subtable}{\textwidth}
        \caption{\small \textit{Sample density=40, train data size=10}}
        \centering
    \begin{tabular}{lccc}
        \toprule
        \textit{Dataset}      & GKN      & Ours    & Error Reduction \\
        \midrule
        Darcy Flow            & 0.6207   & 0.5169  & 16.72\% \\
        Advection             & 1.1147   & 0.6941  & 37.73\% \\
        Eikonal               & 0.2090   & 0.1521  & 27.22\% \\
        Nonlinear Diffusion   & 0.2920   & 0.2439  & 16.47\% \\
        \bottomrule
    \end{tabular}
    
\end{subtable} 

\end{table*}

\begin{table*}

    \begin{subtable}{\textwidth}
        \caption{\small \textit{Sample density=50, train data size=10}}
        \centering
    \begin{tabular}{lccc}
        \toprule
        \textit{Dataset}      & GKN      & Ours    & Error Reduction \\
        \midrule
        Darcy Flow            & 0.5909   & 0.4944  & 16.33\% \\
        Advection             & 1.0714   & 0.6219  & 41.95\% \\
        Eikonal               & 0.2063   & 0.1498  & 27.39\% \\
        Nonlinear Diffusion   & 0.2788   & 0.2048  & 26.54\% \\
        \bottomrule
    \end{tabular}
    
\end{subtable} 

    \begin{subtable}{\textwidth}
        \caption{\small \textit{Sample density=60, train data size=10}}
        \centering
    \begin{tabular}{lccc}
        \toprule
        \textit{Dataset}      & GKN      & Ours    & Error Reduction \\
        \midrule
        Darcy Flow            & 0.5624   & 0.4762  & 15.33\% \\
        Advection             & 1.1333   & 0.6342  & 44.04\% \\
        Eikonal               & 0.2051   & 0.1483  & 27.69\% \\
        Nonlinear Diffusion   & 0.2751   & 0.1963  & 28.64\% \\
        \bottomrule
    \end{tabular}
    
\end{subtable} 

    \begin{subtable}{\textwidth}
        \caption{\small \textit{Sample density=70, train data size=10}}
        \centering
    \begin{tabular}{lccc}
        \toprule
        \textit{Dataset}      & GKN      & Ours    & Error Reduction \\
        \midrule
        Darcy Flow            & 0.5629   & 0.4682  & 16.82\% \\
        Advection             & 1.1024   & 0.6654  & 39.64\% \\
        Eikonal               & 0.2039   & 0.1450  & 28.89\% \\
        Nonlinear Diffusion   & 0.2609   & 0.1776  & 31.93\% \\
        \bottomrule
    \end{tabular}
    
\end{subtable}

    \begin{subtable}{\textwidth}
        \caption{\small \textit{Sample density=80, train data size=10}}
        \centering
    \begin{tabular}{lccc}
        \toprule
        \textit{Dataset}      & GKN      & Ours    & Error Reduction \\
        \midrule
        Darcy Flow            & 0.5795   & 0.4561  & 21.29\% \\
        Advection             & 1.0573   & 0.6365  & 39.80\% \\
        Eikonal               & 0.2041   & 0.1412  & 30.82\% \\
        Nonlinear Diffusion   & 0.2600   & 0.1777  & 31.65\% \\
        \bottomrule
    \end{tabular}
    
\end{subtable}

    \begin{subtable}{\textwidth}
        \caption{\small \textit{Sample density=90, train data size=10}}
        \centering
    \begin{tabular}{lccc}
        \toprule
        \textit{Dataset}      & GKN      & Ours    & Error Reduction \\
        \midrule
        Darcy Flow            & 0.5824   & 0.4362  & 25.10\% \\
        Advection             & 1.1247   & 0.6612  & 41.21\% \\
        Eikonal               & 0.2047   & 0.1401  & 31.56\% \\
        Nonlinear Diffusion   & 0.2597   & 0.1693  & 34.81\% \\
        \bottomrule
    \end{tabular}
    
\end{subtable}

\end{table*}

\end{document}

%% file: emacscomm.tex
\newcommand{\var}{{\rm var}}
\newcommand{\Tr}{^{\rm T}}
\newcommand{\vtrans}[2]{{#1}^{(#2)}}
\newcommand{\kron}{\otimes}
\newcommand{\schur}[2]{({#1} | {#2})}
\newcommand{\schurdet}[2]{\left| ({#1} | {#2}) \right|}
\newcommand{\had}{\circ}
\newcommand{\diag}{{\rm diag}}
\newcommand{\invdiag}{\diag^{-1}}
\newcommand{\rank}{{\rm rank}}
 \newcommand{\expt}[1]{\langle #1 \rangle}
\newcommand{\nullsp}{{\rm null}}
\newcommand{\tr}{{\rm tr}}
\renewcommand{\vec}{{\rm vec}}
\newcommand{\vech}{{\rm vech}}
\renewcommand{\det}[1]{\left| #1 \right|}
\newcommand{\pdet}[1]{\left| #1 \right|_{+}}
\newcommand{\pinv}[1]{#1^{+}}
\newcommand{\erf}{{\rm erf}}
\newcommand{\hypergeom}[2]{{}_{#1}F_{#2}}
\newcommand{\mcal}[1]{\mathcal{#1}}
\newcommand{\bepsilon}{\boldsymbol{\epsilon}}
\newcommand{\brho}{\boldsymbol{\rho}}
\renewcommand{\a}{{\bf a}}
\renewcommand{\b}{{\bf b}}
\renewcommand{\c}{{\bf c}}
\renewcommand{\d}{{\rm d}}  
\newcommand{\e}{{\bf e}}
\newcommand{\f}{{\bf f}}
\newcommand{\g}{{\bf g}}
\newcommand{\h}{{\bf h}}
\newcommand{\bi}{{\bf i}}
\newcommand{\bj}{{\bf j}} 

\renewcommand{\k}{{\bf k}}
\newcommand{\m}{{\bf m}}
\newcommand{\mhat}{{\overline{m}}}
\newcommand{\tm}{{\tilde{m}}}
\newcommand{\n}{{\bf n}}
\renewcommand{\o}{{\bf o}}
\newcommand{\p}{{\bf p}}
\newcommand{\q}{{\bf q}}
\newcommand{\wy}{{\widehat{\y}}}
\newcommand{\wlam}{{\widehat{\lambda}}}
\renewcommand{\r}{{\bf r}}
\newcommand{\s}{{\bf s}}
\renewcommand{\t}{{\bf t}}
\renewcommand{\u}{{\bf u}}
\renewcommand{\v}{{\bf v}}
\newcommand{\w}{{\bf w}}
\newcommand{\x}{{\bf x}}
\newcommand{\y}{{\bf y}}
\newcommand{\z}{{\bf z}}
\newcommand{\A}{{\bf A}}
\newcommand{\B}{{\bf B}}
\newcommand{\C}{{\bf C}}
\newcommand{\D}{{\bf D}}
\newcommand{\E}{{\bf E}}
\newcommand{\F}{{\bf F}}
\newcommand{\G}{{\bf G}}
\newcommand{\Gcal}{{\mathcal{G}}}
\newcommand{\Dcal}{\mathcal{D}}
\newcommand{\Qcal}{{\mathcal{Q}}}
\newcommand{\Pcal}{{\mathcal{P}}}
\newcommand{\Hcal}{{\mathcal{H}}}
\renewcommand{\H}{{\bf H}}
\newcommand{\I}{{\bf I}}
\newcommand{\J}{{\bf J}}
\newcommand{\K}{{\bf K}}
\renewcommand{\L}{{\bf L}}
\newcommand{\Lcal}{{\mathcal{L}}}
\newcommand{\M}{{\bf M}}
\newcommand{\Mcal}{{\mathcal{M}}}
\newcommand{\Ocal}{{\mathcal{O}}}
\newcommand{\Fcal}{{\mathcal{F}}}
\newcommand{\N}{\mathcal{N}}  
\newcommand{\bupeta}{\boldsymbol{\upeta}}
\renewcommand{\O}{{\bf O}}
\renewcommand{\P}{{\bf P}}
\newcommand{\Q}{{\bf Q}}
\newcommand{\R}{{\bf R}}
\renewcommand{\S}{{\bf S}}
\newcommand{\Scal}{{\mathcal{S}}}
\newcommand{\T}{{\bf T}}
\newcommand{\Tcal}{{\mathcal{T}}}
\newcommand{\U}{{\bf U}}
\newcommand{\Ucal}{{\mathcal{U}}}
\newcommand{\tU}{{\tilde{\U}}}
\newcommand{\tUcal}{{\tilde{\Ucal}}}
\newcommand{\V}{{\bf V}}
\newcommand{\W}{{\bf W}}
\newcommand{\Wcal}{{\mathcal{W}}}
\newcommand{\Vcal}{{\mathcal{V}}}
\newcommand{\X}{{\bf X}}
\newcommand{\Xcal}{{\mathcal{X}}}
\newcommand{\Y}{{\bf Y}}
\newcommand{\Ycal}{{\mathcal{Y}}}
\newcommand{\Z}{{\bf Z}}
\newcommand{\Zcal}{{\mathcal{Z}}}

\newcommand{\bfLambda}{\boldsymbol{\Lambda}}

\newcommand{\bsigma}{\boldsymbol{\sigma}}
\newcommand{\balpha}{\boldsymbol{\alpha}}
\newcommand{\bpsi}{\boldsymbol{\psi}}
\newcommand{\bphi}{\boldsymbol{\phi}}
\newcommand{\bPhi}{\boldsymbol{\Phi}}
\newcommand{\bbeta}{\boldsymbol{\beta}}
\newcommand{\Beta}{\boldsymbol{\eta}}
\newcommand{\btau}{\boldsymbol{\tau}}
\newcommand{\bvarphi}{\boldsymbol{\varphi}}
\newcommand{\bzeta}{\boldsymbol{\zeta}}

\newcommand{\blambda}{\boldsymbol{\lambda}}
\newcommand{\bLambda}{\mathbf{\Lambda}}

\newcommand{\btheta}{\boldsymbol{\theta}}
\newcommand{\bpi}{\boldsymbol{\pi}}
\newcommand{\bxi}{\boldsymbol{\xi}}
\newcommand{\bSigma}{\boldsymbol{\Sigma}}
\newcommand{\bPi}{\boldsymbol{\Pi}}
\newcommand{\bOmega}{\boldsymbol{\Omega}}

\newcommand{\bx}{{\bf x}}
\newcommand{\bgamma}{\boldsymbol{\gamma}}
\newcommand{\bGamma}{\boldsymbol{\Gamma}}
\newcommand{\bUpsilon}{\boldsymbol{\Upsilon}}

\newcommand{\bmu}{\boldsymbol{\mu}}
\newcommand{\1}{{\bf 1}}
\newcommand{\0}{{\bf 0}}

\newcommand{\bs}{\backslash}
\newcommand{\ben}{\begin{enumerate}}
\newcommand{\een}{\end{enumerate}}

 \newcommand{\notS}{{\backslash S}}
 \newcommand{\nots}{{\backslash s}}
 \newcommand{\noti}{{\backslash i}}
 \newcommand{\notj}{{\backslash j}}
 \newcommand{\nott}{\backslash t}
 \newcommand{\notone}{{\backslash 1}}
 \newcommand{\nottp}{\backslash t+1}

\newcommand{\notk}{{^{\backslash k}}}
\newcommand{\notij}{{^{\backslash i,j}}}
\newcommand{\notg}{{^{\backslash g}}}
\newcommand{\wnoti}{{_{\w}^{\backslash i}}}
\newcommand{\wnotg}{{_{\w}^{\backslash g}}}
\newcommand{\vnotij}{{_{\v}^{\backslash i,j}}}
\newcommand{\vnotg}{{_{\v}^{\backslash g}}}
\newcommand{\half}{\frac{1}{2}}
\newcommand{\msgb}{m_{t \leftarrow t+1}}
\newcommand{\msgf}{m_{t \rightarrow t+1}}
\newcommand{\msgfp}{m_{t-1 \rightarrow t}}

\newcommand{\proj}[1]{{\rm proj}\negmedspace\left[#1\right]}
\newcommand{\argmin}{\operatornamewithlimits{argmin}}
\newcommand{\argmax}{\operatornamewithlimits{argmax}}

\newcommand{\dif}{\mathrm{d}}
\newcommand{\abs}[1]{\lvert#1\rvert}
\newcommand{\norm}[1]{\lVert#1\rVert}

\newcommand{\mrm}[1]{\mathrm{{#1}}}
\newcommand{\RomanCap}[1]{\MakeUppercase{\romannumeral #1}}
\newcommand{\EE}{\mathbb{E}}
\newcommand{\bbI}{\mathbb{I}}
\newcommand{\bbH}{\mathbb{H}}
\newcommand{\ie}{{\textit{i.e.,}}\xspace}
\newcommand{\eg}{{\textit{e.g.,}}\xspace}
\newcommand{\etc}{{\textit{etc.}}\xspace}
\newcommand{\cmt}[1]{}